\documentclass[journal,twoside]{IEEEtran}

\usepackage{graphicx}
\usepackage{wrapfig}
\usepackage{colortbl}
\usepackage{float}
\usepackage{overpic}
\usepackage{multirow}
\usepackage{booktabs}
\usepackage{bm}
\usepackage{verbatim}
\usepackage{color}
\usepackage{arydshln}
\usepackage{makecell,rotating}
\usepackage{array}
\usepackage{amsmath}
\usepackage{amssymb}
\usepackage{amsthm}
\usepackage{graphicx}
\usepackage{subfigure}
\usepackage{cite}
\usepackage{color}
\usepackage{epstopdf}
\usepackage{subeqnarray}
\usepackage{cases}
\usepackage{amsthm,amsmath,amssymb}
\usepackage{mathrsfs}
\usepackage{multirow}
\usepackage{algorithm}
\usepackage{algorithmicx}
\usepackage{algpseudocode}
\floatname{algorithm}{Algorithm}

\usepackage[dvipsnames]{xcolor}
\DeclareMathAlphabet{\mathscr}{OT1}{pzc}{m}{it}
\usepackage{enumitem}
\setenumerate[1]{itemsep=4pt,partopsep=0pt,parsep=\parskip,topsep=4pt}
\setitemize[1]{itemsep=4pt,partopsep=0pt,parsep=\parskip,topsep=4pt}

\definecolor{citecolor}{RGB}{119,185,0}
\definecolor{citecolor1}{RGB}{66,168,235}
\usepackage{hyperref}
\hypersetup{colorlinks=true,citecolor=citecolor1,linkcolor=BrickRed,urlcolor=Thistle}

\definecolor{mygray}{gray}{.88}
\definecolor{mygrayd}{gray}{.94}

\def\eg{\emph{e.g. }}
\def\etc{\emph{etc}}
\def\etal{{\em et al.~}}

\begin{document}

\title{Location-Free Camouflage Generation Network}

\author{Yangyang Li$^{\textbf{*}}$, Wei Zhai$^{\textbf{*}}$, Yang Cao, \IEEEmembership{Member, IEEE}, and Zheng-jun Zha, \IEEEmembership{Member, IEEE}

\thanks{$\textbf{*}$ Yangyang Li and Wei Zhai contributed equally to this paper.}

\thanks{Yangyang Li, Wei Zhai, Yang Cao, Zheng-jun Zha are with the School of Information Science and Technology, at the University of Science and Technology of China, Anhui, China. (email: \{lyy1030, wzhai056\}@mail.ustc.edu.cn, \{forrest, zhazj\}@ustc.edu.cn).}
}

\maketitle
\begin{abstract}
Camouflage is a common visual phenomenon, which refers to hiding the foreground objects into the background images, making them briefly invisible to the human eye. Previous work has typically been implemented by an iterative optimization process. However, these methods struggle in 1) efficiently generating camouflage images using foreground and background with arbitrary structure; 2) camouflaging foreground objects to regions with multiple appearances (\eg the junction of the vegetation and the mountains), which limit their practical application. To address these problems, this paper proposes a novel {\itshape \textbf{L}ocation-free \textbf{C}amouflage \textbf{G}eneration \textbf{Net}work} (\textbf{LCG-Net}) that fuse high-level features of foreground and background image, and generate result by one inference. Specifically, a {\itshape Position-aligned Structure Fusion} (PSF) module is devised to guide structure feature fusion based on the point-to-point structure similarity of foreground and background, and introduce local appearance features point-by-point. To retain the necessary identifiable features, a new immerse loss is adopted under our pipeline, while a background patch appearance loss is utilized to ensure that the hidden objects look continuous and natural at regions with multiple appearances. Experiments show that our method has results as satisfactory as state-of-the-art in the single-appearance regions and are less likely to be completely invisible, but far exceed the quality of the state-of-the-art in the multi-appearance regions. Moreover, our method is hundreds of times faster than previous methods. Benefitting from the unique advantages of our method, we provide some downstream applications for camouflage generation, which show its potential. The related code and dataset will be released at \url{https://github.com/Tale17/LCG-Net}.
\end{abstract}


\section{Introduction}
\IEEEPARstart{C}{amouflage} images \cite{chu2010camouflage} are both a wide range of visual phenomena in the natural world and an artistic form in human society. These images aim to hide the foreground objects somewhere in the backgrounds and to ensure that the overall structures and appearances are still natural and continuous, thus it is hard for the viewers to spot them at a glance, unless through more detailed observation. This technology is widely used in artworks \cite{steven2010, kim2019style}, visual effects, and generating datasets for camouflage object detection \cite{Camouflage20CVPR, 21PAMI-Concealed}. The challenge of this task is that the embedding parts must have a continuous structure and appearance with surrounding regions, otherwise, it will cause the foreground objects to be too standout. In addition, the foreground objects must still be identifiable. Therefore, simply using style transfer\cite{gatys2016image} or texture synthesis\cite{gatys2015texture} will get unnatural results, since these methods replace global appearances such as style or texture while maintaining the whole foreground content, ignoring both the content conflict between the two images and the distinction between different appearances.


Recognizing camouflage images is challenging for humans. According to the feature integration theory \cite{treisman1980feature}, the human visual perceptual mechanism is divided into two main phases, namely {\itshape  feature search} and {\itshape  conjunction search} \cite{treisman1988features,wolfe1994guided}. Feature search is a relatively fast process, which perceives objects by their intuitive features such as color, texture, edge, \etc. On the contrary, conjunction search is slower. It recognizes objects by integrating features scattered in different spatial locations. Camouflage images mostly use the features in the background which support fast search to replace the corresponding features of the foreground objects, so humans cannot identify the objects at first, but can gradually find the clues retained for conjunction search and complete the identification.

\begin{figure}[t]
  \centering
  \small
		\begin{overpic}[width=1.\linewidth]{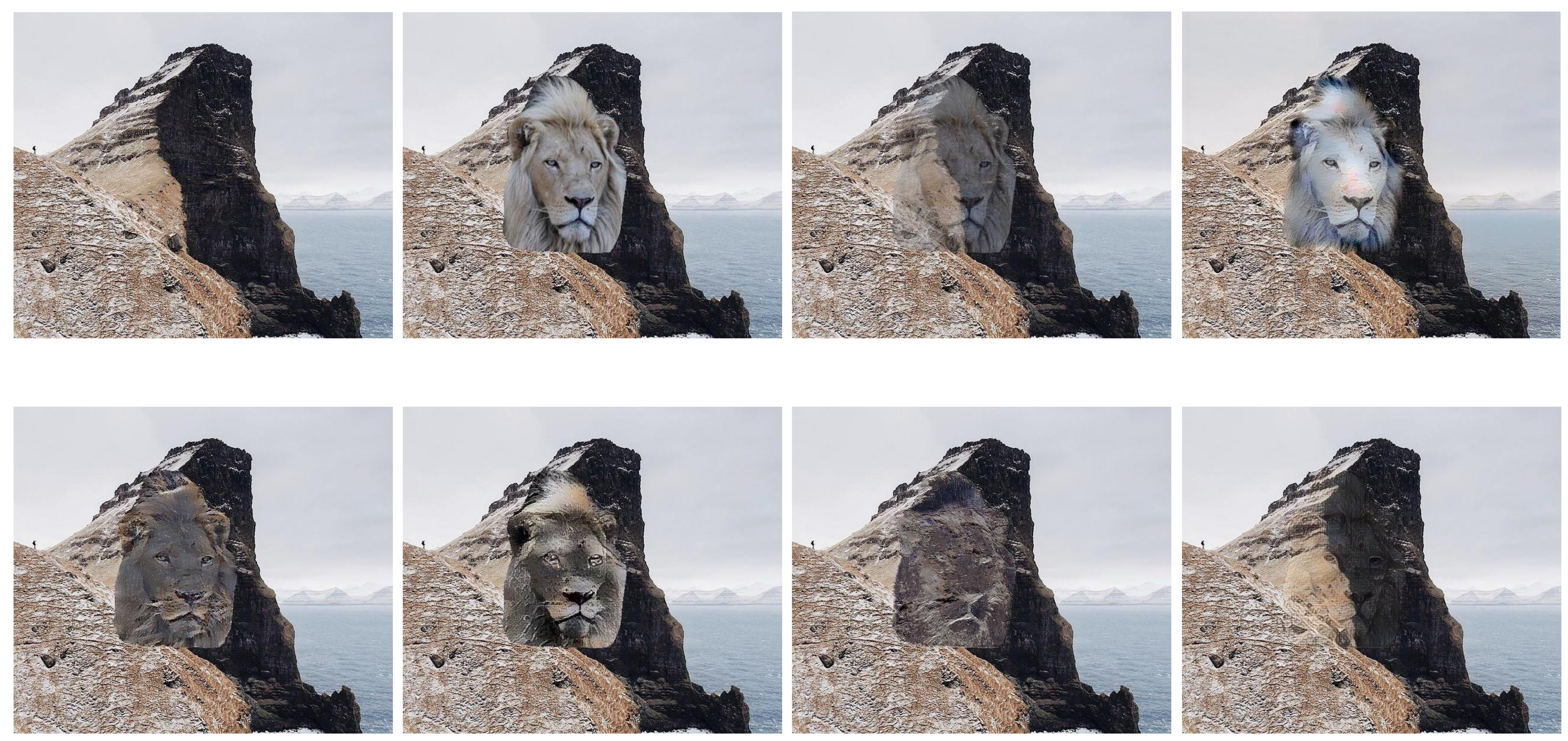}
		\put(4, 22.5){\textbf{Background}}
		\put(30, 22.5){\textbf{Embeding}}
		\put(57, 22.5){\textbf{AB~\cite{perez2003poisson}}}
		\put(81, 22.5){\textbf{DIB~\cite{zhang2020deepBlend}}}
		\put(9, -3.){\textbf{NS~\cite{gatys2016image}}}
		\put(29, -3.){\textbf{AdaIN~\cite{huang2017arbitrary}}}
		\put(50, -3.){\textbf{Zhang~\etal \cite{zhang2020deep}}} 
		\put(83, -3.){\textbf{Ours}}
		\end{overpic}
  \caption{Overview of the comparison of our method with existing methods in the region with multiple appearances. Our method has the best visual effect.}
  \label{fig:fig1}
\end{figure}
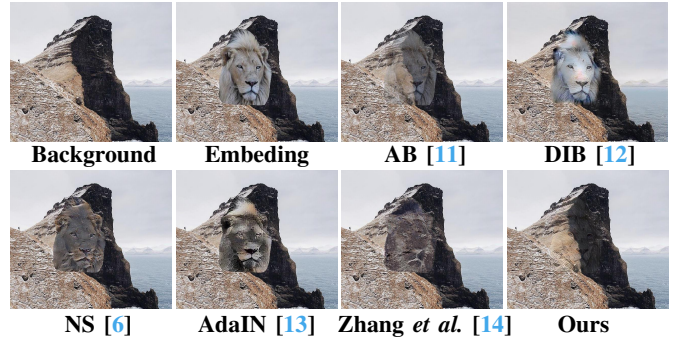

Recently, there are few methods to generate camouflage images. Chu \etal \cite{chu2010camouflage} propose the first camouflage method based on hand-crafted features. They optimize the luminance and texture of the foreground objects according to the feature integration theory \cite{treisman1980feature}. But their method may fail when the luminance contrast of the foreground is fairly low. Zhang \etal \cite{zhang2020deep} propose the first deep camouflage method based on iterative optimization. They design an attention-aware camouflage loss to only retain the salient features of foreground objects. However, their method is based on a slow iterative optimization, which limits their practical application. And they refer to style transfer \cite{gatys2016image} and introduce the global style of the backgrounds for foreground objects, lacks the distinction of different local spatial features in the backgrounds, thus creating an obvious boundary in regions with multiple appearances (\eg across mountains and sky). 

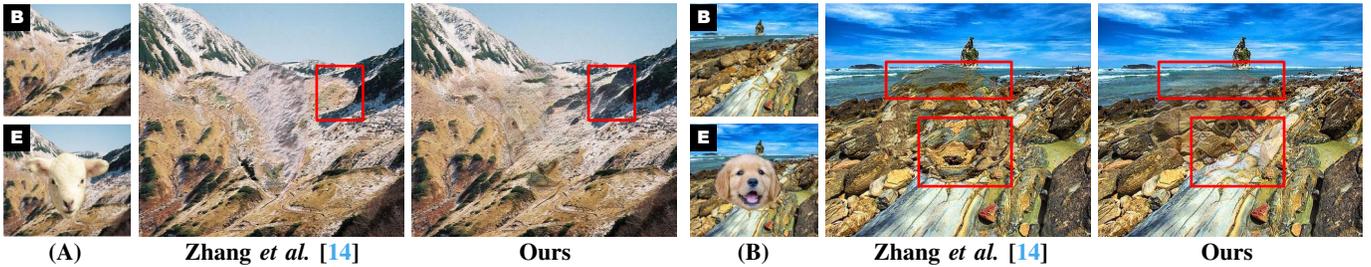
\begin{figure*}[t]
	\centering
	\small
		\begin{overpic}[width=1.\linewidth]{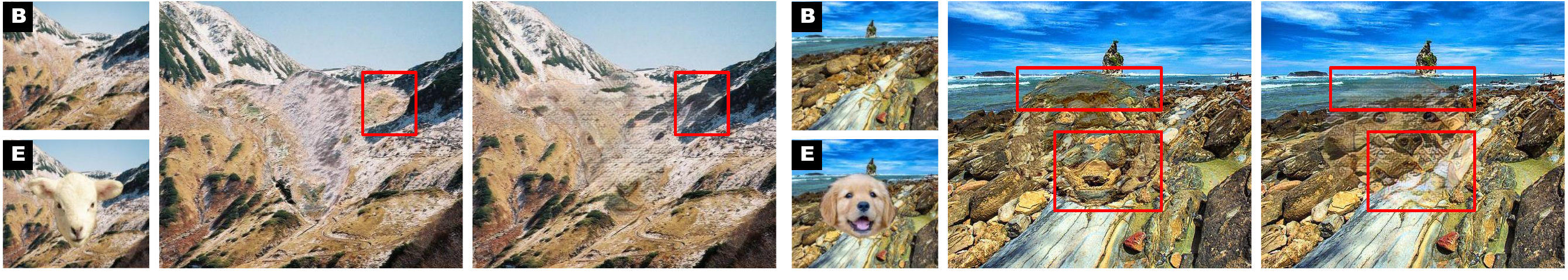}
		\put(13.5, -1){\textbf{Zhang~\etal \cite{zhang2020deep}}}
		\put(38, -1){\textbf{Ours}}
		\put(64, -1){\textbf{Zhang~\etal \cite{zhang2020deep}}}
		\put(88, -1){\textbf{Ours}}
        \put(3.5,-1){\textbf{(A)}}
		\put(54,-1){\textbf{(B)}}
		\end{overpic}
	\caption{Comparison of the SOTA method \cite{zhang2020deep} and ours in regions with complex structures and multiple appearances. Part (A): Structure consistency. They break the continuation of vegetation at the marked position, while we preserve this structure and are therefore more visually natural.  Part (B): Appearance consistency. They mix multiple appearances of the camouflage region, making the foreground too standout, while we distinguish the various appearances and achieve the purpose of camouflage.}
	\label{consist}
\end{figure*}

Analyzing the dilemma of previous work, to improve the adaptability of camouflage generation to diverse backgrounds, except for retaining the necessary information to identify foreground objects, the following two points should be followed: \textbf{Structure consistency} and \textbf{Appearance consistency}. \textbf{Structure Consistency} requires that the output has a continuous structure, so the embedded part should destroy the background structure of the embedded region as little as possible (Fig. \ref{consist} (A)). Therefore, we need to choose a set of foreground structure features that have sufficient characterization capabilities and can be harmoniously fused with the background structure. \textbf{Appearance Consistency} requires the embedded part to have similar local appearance features (\eg color and texture) to the embedded regions. These features can neither be mixed together nor randomly distributed, otherwise, it is easy to cause visual discontinuity (Fig. \ref{consist} (B)). Therefore, these appearance features should be spatially aligned with the embedding region. As mentioned above, style transfer only retains the foreground content and does not consider background structure, so is difficult to obtain a coherent structure. Zhang \etal \cite{zhang2020deep} select structure features which can characterize foreground objects according to their saliency maps, but the preserved foreground parts are only decided by foregrounds, so is difficult to adapt to diverse backgrounds.  And both two methods are not suitable for multi-appearance regions since they replace the foreground style with the background global style. 

To address these issues, we propose the first arbitrary image camouflage pipeline, named location-free camouflage generation network (LCG-Net), which efficiently camouflage the foreground objects to any region with single or multiple appearances. Specifically, we adopt a novel {\itshape Position-aligned Structure Fusion} (PSF) module, which adjusts the fusion ratio of the foreground and background structure features of each point according to their point-to-point structure similarity in the spatial dimension, and introduces the local appearance features of the corresponding position on the background features for each point. PSF module preserves different foreground structure features for backgrounds with different structures, thus having good adaptability to varying backgrounds. Based on PSF module, a novel background patch appearance (BPA) loss is adopted to constrain the local appearance (from a small neighborhood) of each point to be similar to the background. And a new immerse (IM) loss is utilized to calculate structure loss more intuitively.

Our contributions are summarized as follows:
\begin{itemize}

\item [1)]
We introduce a data-driven framework named {\itshape \textbf{L}ocation-free \textbf{C}amouflage \textbf{G}eneration \textbf{Net}work} (\textbf{LCG-Net}) for camouflage generation, which only needs once inference to get the camouflage image and is hundreds of times faster than previous methods.

\item [2)]
We present a novel PSF module, which adaptively fuses the foreground and background features by point-to-point structure similarity and introduces the local appearance features of the background for each point, thus the output has a consistent structure and appearance. 

\item [3)]
We design a new BPA loss to accurately measure the appearance difference in multi-appearance regions, and a new IM loss is adopted to more intuitively retain necessary structure clues for identification.

\item [4)]
Experiments and user studies show that our method has an equivalent effect in the single-appearance regions to the SOTA, while is significantly better than the SOTA in the multi-appearance regions. And we enumerate some practical applications for camouflage generation.

\end{itemize}

The rest of the paper is organized as follows. Section \ref{related} illustrates the previous work related to camouflage generation. The novel proposed LCG-Net and PSF module and our loss functions are introduced in Section \ref{method}. Section \ref{exp} shows the experiment results and analysis based on ECG-Net. Section \ref{app} gives some potential applications of camouflage generation. Section \ref{limitation} analyzes the limitations of our approach and some possible workarounds. We conclude the paper by summarizing and future work in Section \ref{conclusion}.

\section{Related Work}
\label{related}
Camouflage images are designed to seamlessly embed foreground objects into the background images, making them briefly invisible to humans. Tasks such as style transfer, texture synthesis, and image blending also fuse two images under certain constraints. Therefore, we introduce these tasks in this section and explain their shortcomings when applied to camouflage generation at the end of each subsection. 

\subsection{Texture Synthesis \& Style Transfer}
Texture synthesis requires that the fragments before and after synthesis are similar under human cognition, and have no discontinuous fragments, which is similar to the appearance consistency. Heeger \etal \cite{heeger1995pyramid} achieve this goal through histogram matching on linear filter responses. Efros \etal \cite{efros1999texture} use a non-parametric sampling method based on field filling to achieve a better effect, but this method is too slow. Therefore, Efros \etal \cite{efros2001image} propose another faster texture synthesis method based on image quilting. Gatys \etal \cite{gatys2015texture} use CNN to synthesize textures by matching correlations in the feature space, and it also inspires the later style transfer work. Lockerman \etal \cite{lockerman2016multi} guide the texture synthesis process by using extra guidance channels to preserve the large-scale structure in regular or near-regular textures. Li \etal \cite{li2018non} captures the long-range structure of a texture based on non-local operators without the additional guidance channel. However, only using texture synthesis to camouflage the objects lacks the reconciliation for the foreground and background structure, therefore the structural naturalness of the whole image may be seriously damaged and cause the camouflage to fail.

Style transfer is to keep the content of the original image while introducing the style of the target image. Some early methods \cite{efros2001image, frigo2016split} achieved this goal by matching low-level statistics. But due to the lack of understanding of semantics, the results are often unsatisfactory. Gatys \etal \cite{gatys2016image} obtain an impressive style transfer effect through matching high-level feature statistics of images for the first time. However, their method is based on a slow iterative optimization process, so the migration time is longer. Some methods \cite{johnson2016perceptual, ulyanov2016texture, li2016precomputed, ulyanov2017improved, wang2017multimodal} based on feed-forward networks solve this problem, but can only match a few styles. Li \etal \cite{li2017universal} design a multi-style transfer network by instance regularization. Huang \etal \cite{huang2017arbitrary} propose a feed-forward network that can achieve arbitrary style transfer based on adaptive instance regularization so that the number of styles and the transfer speed are no longer contradictory. After that, a series of arbitrary style transfer methods \cite{jing2020dynamic, li2018learning, deng2020arbitrary, chen2016fast, sheng2018avatar, park2019arbitrary, liu2021adaattn} are proposed. Virtusio \etal \cite{virtusio2021neural} present a method for interactively generating a variety of stylized images from only a single style input. Based on arbitrary style transfer methods, some video style transfer methods \cite{gupta2017characterizing, chen2017coherent, huang2017real} with time consistency have emerged. Similar to texture synthesis, style transfer does not pay attention to structure information of the background, so it will also fail in camouflage since destroying the structure consistency of the output. And style transfer cannot distinguish different styles in the image, so it will cause serious style aliasing when the background image has multiple styles. 

Unlike the above two tasks, we guide fusion by the structure similarity of foreground and background, and introduce local appearance at each point to prevent style aliasing, therefore is competent for camouflage generation. 

\subsection{Image Blending}
Image blending shares some similarities with camouflage generation, which aims to seamlessly add an object from a source image onto a target image, making it look like the object is present in the original scene. The earliest method uses alpha blending that blends at a specified ratio, but it produces a ghost effect in the blending region. Some subsequent methods are dedicated to reconstructing the pixels of the mixed region and ensuring the consistency of its gradient-domain with the original image \cite{perez2003poisson, agarwala2004interactive, jia2006drag, levin2004seamless, uyttendaele2001eliminating, fattal2002gradient, kazhdan2008streaming}. On this basis, Zhang \etal \cite{zhang2020deepBlend} present a Poisson blending loss and jointly train it with style and content loss to reconstruct the blending region. Luan \etal \cite{luan2018deep} propose a two-stage optimization method to achieve coarse and fine coordination. Xiao \etal \cite{xiao2019multi} propose a multi-focus image fusion algorithm based on Hessian matrix. However, these methods only pay attention to the harmony of the embedded objects with the surroundings and do not require them to be temporarily invisible, which runs counter to camouflage generation.


\subsection{Camouflage Generation}
There is a little work dedicate to camouflage generation. Chu \etal \cite{chu2010camouflage} propose the first work specifically to address the camouflage generation. Based on the feature integration theory \cite{treisman1988features, wolfe1994guided}, they optimize hand-crafted features and embed hidden objects in the background images. Although their method can generate attractive effects in some situations, it is sensitive to luminance differences of hidden objects since they rely on the luminance assignment of the foreground. Inspired by neural style transfer \cite{gatys2016image}, Zhang \etal \cite{zhang2020deep} propose a deep camouflage generation method. They introduce an attention-aware camouflage loss to selectively retain information of hidden objects at the feature level and achieved satisfactory effects. However, they select features to keep based on foreground saliency only, lacking adaptability when facing different backgrounds. And they use style transfer \cite{gatys2016image} to transfer appearance features in the backgrounds, thus also cannot adapt to multi-appearance regions. Zhang \etal \cite{zhang2016reversible} hide reversible data in encrypted images based on reversible image transformation, but their goal is to hide encrypted information and restore it according to a specific key, not a visual phenomenon. Base on the results obtained from the camouflage generation, some other work such as breaking \cite{tankus2001convexity, reynolds2011interactive} or detecting \cite{fan2021concealed, mei2021camouflaged, zhai2021mutual} camouflage images are extended. 


\begin{figure*}[t]
  \centering
  \begin{overpic}[width=1.\linewidth]{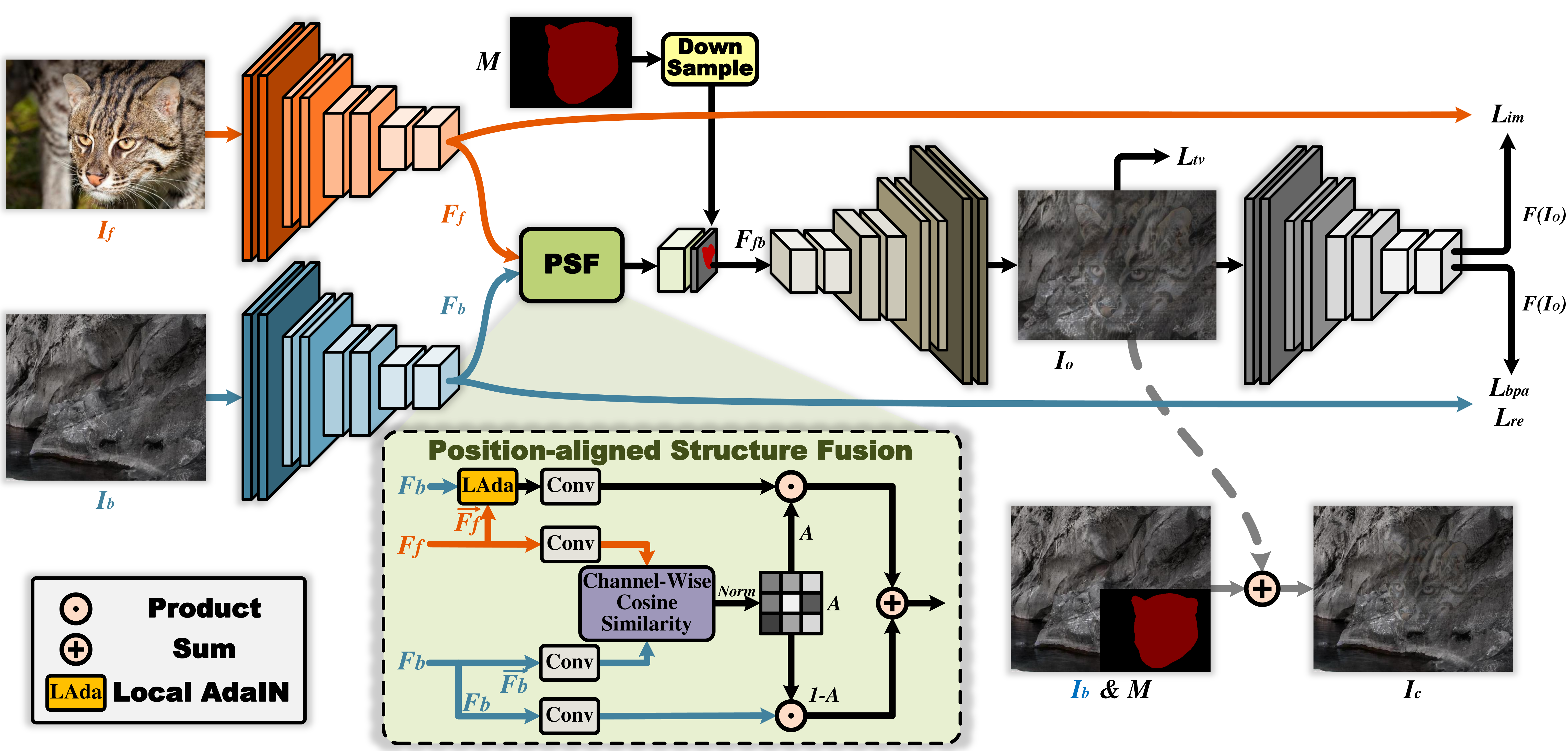}
		\put(51,42){\textbf{Section} (\ref{LCG})}
		\put(24.7,9){\textbf{Section} (\ref{PSF})}
		\put(72,38.5){(\textbf{Eq.} \ref{eq12})}
		\put(92.5,42){(\textbf{Eq.} \ref{eq9})}
		\put(92.5,19.){(\textbf{Eq.} \ref{eq10})}
		\put(87,23.5){(\textbf{Eq.} \ref{eq11})}
		\put(77.7,7.5){(\textbf{Eq.} \ref{eq3})}
  \end{overpic}
  \caption{Overview of Location-free Camouflage Generation Network (LCG-Net). LCG-Net accepts foreground image $I_f$ and background image $I_b$ with the same size. The encoder is taken from the first four layers of VGG-19 for extracting the high-level features. PSF module fuses the foreground and background features and concatenates the result with the downsampled mask at the channel dimension to obtain $F_{fb}$. $F_{fb}$ is fed to the decoder to generate the output image $I_o$. We finally embed $I_o$ into $I_b$ to get the refined result $I_c$. We use the same structure as the Encoder $E$ to calculate the loss functions. The Position-aligned Structure Fusion (PSF) module adaptively fuses the high-level features of the foreground and background according to their point-to-point structure similarity. Norm represents the normalization operation. $\overline{F}$ denotes the channel-wise mean-variance normalized version of $F$.}
  \label{fig:framework}
\end{figure*}
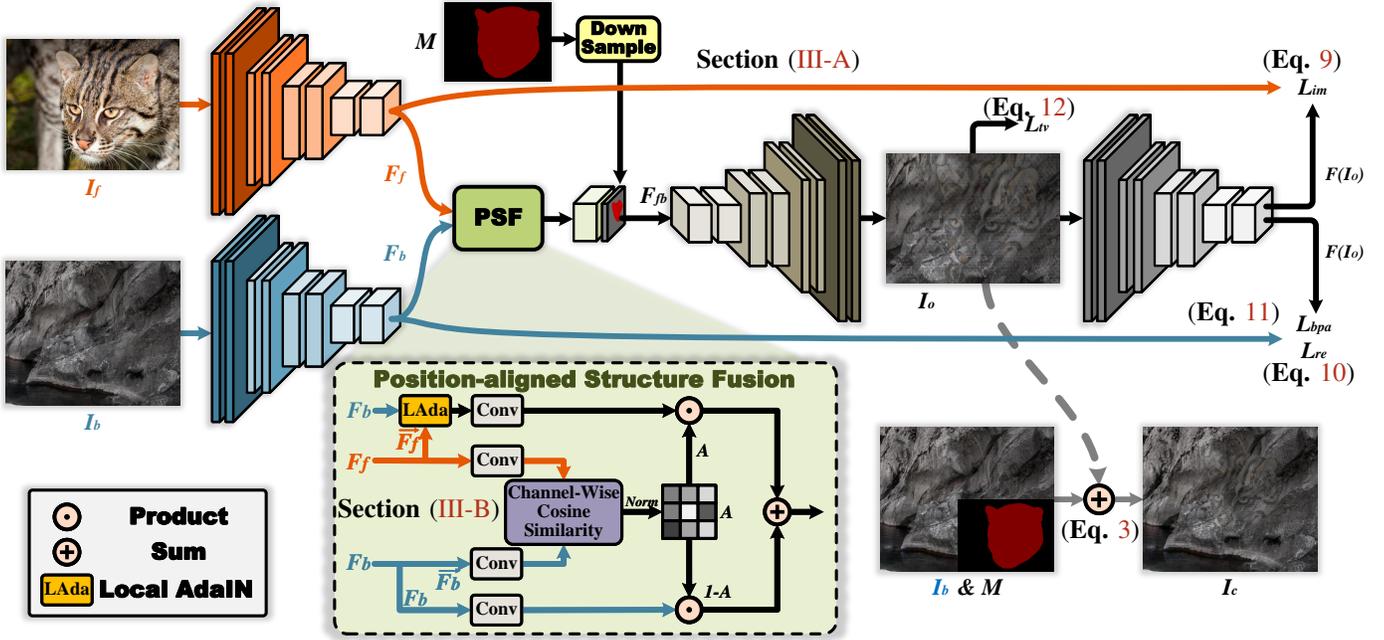

\section{Method}
\label{method}
As shown in Fig. \ref{fig:framework}, our approach receives a foreground image containing a hidden object and a background image of the same size, efficiently generates a camouflage image, and can still get high-quality results in multi-appearance or structure-complex regions, thus is location-free. To get the two images of the same size, we simply crop a region the same size as the foreground image from any position in the original background image that we want to camouflage to (the foreground image is smaller than the background image in most cases). To simplify the description, the background image mentioned below refers to the part cropped to the same size as the foreground image. Since we do not want the background region outside the mask indication to change during the camouflage process, and the codec-based model cannot reconstruct the background image without any errors, we add an embedding operation after LCG-Net to get more refined results. In this section, we assume that we only camouflage one foreground object into one background image. Multi-objects camouflage in the same background can be achieved by camouflaging each object into different regions in turn. Our approach does not modify the region outside the current camouflage region of the background image, so the camouflage objects added later will not affect the previous camouflage effect (as shown in Fig. \ref{fig:multi_object}).

\subsection{Location-free Camouflage Generation Network}
\label{LCG}
To generate camouflage images at a faster speed and improve the adaptation to different backgrounds, we design a {\itshape \textbf{L}ocation-free \textbf{C}amouflage \textbf{G}eneration \textbf{Net}work} (\textbf{LCG-Net}). LCG-Net consists of three components: an encoder $E$ for extracting high-level features of foreground image $I_f$ and background image $I_b$, a decoder $D$ for reconstructing images from the feature space, and an {\itshape position-aligned structure fusion} (PSF) module for fusing features extracted from $I_f$ and $I_b$ according to their local structure similarity.

Specifically, LCG-Net receives a foreground image $I_f$ and a background image $I_b$ of the same size, and a shared mask $M$ as inputs, $M$ indicates the object to be camouflaged in $I_f$ and the camouflage region (can be selected arbitrarily) in $I_b$. The encoder is taken from the first four layers of a pre-trained Vgg-19 \cite{simonyan2014very} (until ReLU 4-1), $I_f$ and $I_b$ are encoded into the feature space and obtain $F_f = E(I_f)$, $F_b = E(I_b)$, where $F$ is a feature map of the shape $C \times W \times H$. The PSF module adaptively fuses $F_f$ and $F_b$ according to their point-to-point structure similarity and produces the output feature map:
\begin{equation}
\label{eq1}
  F_{fb} = Cat(PSF(F_f, F_b), M_d),
\end{equation}
where $M_d$ represents $M$ after downsampling, $Cat$ represents the concatenate operation at the channel dimension. We concatenate $M_d$ in the fused feature map to guide the decoder to pay more attention to the part indicated by $M$. Then $F_{fb}$ is sent to the decoder D to get the output of LCG-Net:
\begin{equation}
\label{eq2}
  I_o = D(F_{fb}).
\end{equation}
The decoder is the only part of our network which needs to be trained, it is a mirror structure with the encoder, and is used to reconstruct the feature map to the original image size. Referring to the work of Huang \etal \cite{huang2017arbitrary}, all pooling layers in the decoder are replaced with nearest up-sampling to reduce checkerboard effects. Refer to \cite{ulyanov2016texture, ulyanov2017improved, dumoulin2016learned}, we feed the output image $I_o$ to the same pre-trained VGG-19 as the encoder to compute the loss functions and optimize the decoder. After the training process, we can generate $I_o$ for any $I_f$ and $I_b$ through one forward propagation, which greatly increases the generation speed.

Please note that $I_o$ is not our final result. As mentioned at the beginning of this section, $I_o$ generated by the codec-based model is not completely consistent with $I_b$ in the region outside $M$. Therefore, we embed the camouflaged object indicates by $M$ in $I_o$ into $I_b$: 
\begin{equation}
\label{eq3}
  I_c = I_o \otimes M + I_b \otimes (1 - M),
\end{equation}
where $\otimes$ represents the matrix dot product.


\subsection{Position-aligned Structure Fusion Module}
\label{PSF} 
For appearance consistency in multi-appearance regions, a conventional idea is to extract features for each different region based on semantics in $I_b$ and use them to camouflage the corresponding region in $I_f$. But this approach will greatly increase the algorithm complexity. Zhang \etal \cite{zhang2020deep} escape from the appearance consistency, they recommend a region with a consistent appearance as the camouflage region, but this approach greatly limits the practicality of camouflage generation since its restricted camouflage region. The issue of structure consistency has not been well resolved too. Most approaches do not involve the fusion of foreground and background at the structure level. Although Zhang \etal \cite{zhang2020deep} fuse features based on saliency maps of foreground image and uses background features filling the less salient foreground region, the preserved background structure is only determined by the saliency of the foreground, therefore hard to ensure that the background structure is not damaged.

Based on the dilemma of the predecessors, we propose the {\itshape Position-aligned Structure Fusion (PSF)} module. As shown in Fig. \ref{fig:framework}, the PSF module fuses $F_f$ and $F_b$ according to their point-to-point structure similarity and introduces the local appearance feature of the background corresponding position for each point, thus obtaining an output with structure consistency and appearance consistency. Since the relative positions of the $I_f$ and $I_b$ have been determined, this point-to-point contrast is necessary. The more similar the foreground and the background structure, the more foreground features are retained for identification, otherwise, more background features are retained to protect the background structure from being destroyed.

Specifically, we first calculate the point-to-point structure similarity coefficient matrix $A$ of the foreground structure and the background structure: 
\begin{equation}
\label{eq4}
A = Norm(Cos(f(\overline{F_f}),g(\overline{F_b}))),
\end{equation}
where $f$, $g$ are $1\times1$ learnable convolution layers, $\overline{F}$ denotes a channel-wise mean-variance normalized version of $F$. According to Huang \etal \cite{huang2017arbitrary}, this operation can remove the appearance features in $F_f$ and $F_b$, and only retain the structure features to be compared. $Cos$ represents the channel-wise cosine distance similarity, by which we measure the structure similarity of each point pair. $Norm$ denotes a min-max scaling to scale the result to the interval (0, 1). 

For a point on $A$, the larger the value of the point, the more similar the foreground and background structure corresponding to this point. Therefore, we use $A$ as the foreground coefficient matrix and $1-A$ as the background coefficient matrix for fusing the background features and the foreground features after introducing local appearance features: 
\begin{equation}
\label{eq5}
PSF(F_f,F_b) = A \otimes e(LAda(\overline{F_f},F_b)) + (1 - A) \otimes h(F_b),
\end{equation}
where $e$, $h$ are $1\times1$ learnable convolution layers, $LAda$ means local AdaIN, which refers to AdaIN \cite{huang2017arbitrary} to introduce the local mean and variance of the background corresponding position to each point in $\overline{F_f}$ as the appearance features. $LAda$ is obtained by the following equation: 
\begin{equation}
\label{eq6}
LAda(\overline{F_f},F_b)_i = \sigma(F_{b_i}^{\omega\times\omega}) (\overline{F_{f_i}} + \mu(F_{b_i}^{\omega\times\omega})),
\end{equation}
where $i$ denotes the position index, we do this operation for each position. $\mu$ denotes mean and $\sigma$ denotes variance. $F_i^{\omega\times\omega}$ denotes a window of size $\omega\times\omega$ and centered on $F_i$, $\omega$ is set to 7 by default.  For each point $\overline{F_f}_i$ in $\overline{F_f}$,  $LAda$ extracts the appearance features from a fixed-size window around the corresponding point of the background and introduces it into $\overline{F_f}_i$. Since the window size is small, most of the windows fall in the region with a single appearance, and for those windows that contain regions with multi-appearance, each will only affect one point on the foreground feature map, so it can be approximated that $LAda(\overline{F_f}, F_b)$ has the same appearance features as the background, thus the result of Eq. \ref{eq5} also has an appearance consistent with the background in everywhere.

In short, since the $1 \times 1$ convolutions used are all learnable, the PSF module can learn the similarity between the foreground structure and the background structure, and use this as a coefficient matrix to guide the fusion. And PSF module uses local AdaIN to introduce local appearance features from the background, making the fused feature map meets both structure consistency and appearance consistency.

\subsection{Loss Functions}
Inspired by style transfer \cite{gatys2016image, huang2017arbitrary} and Zhang \etal \cite{zhang2020deep}, we define a total loss function that generates $I_o$ for the $I_f$ and $I_b$ according to $F_{fb}$. It can be written as: 
\begin{equation}
\label{eq7}
  \mathcal{L} = \lambda_{im}\mathcal{L}_{im} + \lambda_{re}\mathcal{L}_{re} + \lambda_{bpa}\mathcal{L}_{bpa} + \lambda_{tv}\mathcal{L}_{tv},
\end{equation}
where $\mathcal{L}_{im}$ is immerse loss, $\mathcal{L}_{re}$ is remove loss, $\mathcal{L}_{bpa}$ is background patch appearance loss, $\mathcal{L}_{tv}$ is total variational loss, $\lambda_{im}$, $\lambda_{re}$, $\lambda_{bpa}$, $\lambda_{tv}$ are corresponding weights. We feed $I_o$ to the VGG-19 and obtain a feature map $F_o$ to calculate $\mathcal{L}_{im}$, $\mathcal{L}_{re}$ and $\mathcal{L}_{bpa}$ on feature-level, while $\mathcal{L}_{tv}$ is defined on the pixel-level for $I_o$.

\begin{figure}[t]
  \centering
  \includegraphics[width=1\linewidth]{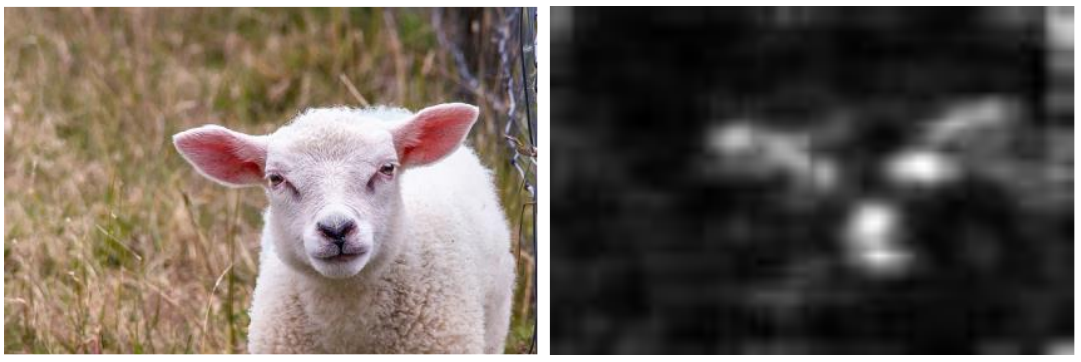}
  \caption{Saliency map of the foreground image. It focuses on eyes, snout, and other features that help us to identify the sheep. Some unexpected noises appear at the edges of the saliency map but do not affect the camouflage, since the objects do not appear at the edges of the image in most cases.}
  \label{fig:saliency}
\end{figure}

\textbf{Immerse Loss.} 
To ensure that the foreground object is still identifiable, the camouflage image needs to retain enough foreground object information for identification. But for structure consistency and appearance consistency, this part of information must be as little as possible. To strike a balance, we design an immerse (IM) loss to compare the contrast difference between $F_o = E(I_o)$ and $F_f$ in the region of our concern. Specifically, we first use the method of Hou \etal \cite{hou2007saliency} to generate a saliency map $S$ (as shown in Fig. \ref{fig:saliency}) for $F_f$: 
\begin{equation}
\label{eq8}
  S = \sum_{k}\left| \mathcal{F}^{-1}[exp\{L(f_k) - L(f_k) \times h_n + iP(f_k)\}] \right|^2,
\end{equation}
where k indexes the feature channels, $\mathcal{F}^{-1}$ is Inverse Fourier Transform, $h_n$ is a local average filter of size $n \times n$. $L$ and $P$ represent the real and imaginary parts of Fourier Transform, $i$ is the imaginary unit. The saliency map $S$ represents the importance of each point on $F_f$ for identifying the object. The larger the value, the more important the point is, so it is more necessary to be retained. We also introduce $M_d$ to select those points that we focus on. Based on $S$ and $M_d$, IM loss calculates the contrast of each point pair on $\overline{F_o}$ and $\overline{F_f}$ respectively, and then calculates the difference of the corresponding contrast on the two feature maps:
\begin{equation}
\label{eq9}
 \begin{split}
 \mathcal{L}_{im} = &\sum_i\sum_{j\neq i}\left|\left|((\overline{F_{o_i}}-\overline{F_{o_j}}) - (\overline{F_{f_i}}-\overline{F_{f_j}}))\right|\right|_2 \\
 &\cdot (S_i+S_j)\cdot (M_{d_i}\lor M_{d_j}),
 \end{split}
\end{equation}
where $i, j$ denote the position index, $\lor$ denotes logical {\itshape OR}, those point pairs with any point on $M_d$ is 1 can be included in our focus. 

IM loss and the leave loss of Zhang \etal \cite{zhang2020deep} both use a saliency map to guide the foreground features that need to be retained, but the difference is that they compute the normalized cosine distance between feature vectors of an input image, which cannot intuitively reflect the structural difference, and the calculation is complicated. Our IM loss directly compares the contrast difference of the feature map after the mean-variance normalization. This contrast reflects the structure of the image, and the calculation is simpler. 

\textbf{Remove Loss.} Using IM loss alone is not sufficent, because it only retains the key structure of the foreground and does not compare with the background structure, which results in structure inconsistency with the background. To address this problem, we use the remove loss of Zhang \etal \cite{zhang2020deep} to introduce structures similar to the background outside the region indicated by $S$, which is defined as: 
\begin{equation}
\label{eq10}
\mathcal{L}_{re} = \sum_{\mathscr{l}=1}^4\left|\left| (1-S) \odot (E^\mathscr{l}(I_o) - E^\mathscr{l}(I_b)) \right|\right|_2,
\end{equation}
where $E^\mathscr{l}(\cdot)$ denotes the feature of VGG encoder in the $\mathscr{l}$-th layer. In our experiments we use $relu1\_1$, $relu2\_1$, $relu3\_1$, $relu4\_1$ layers with equal weights to reconstruct the background structure at multiple scales. 

Zhang \etal \cite{zhang2020deep} introduces remove loss to remove foreground features that support feature search, without considering whether the introduced background structure looks natural. Different from them, by the influence of the PSF module, those foreground points whose structure is too different from the current background point will not be sent to the decoder, so even if the $S$ value of a point is large, it will not destroy the background structure. Under the combined action of IM loss, remove loss, and the PSF module, we can reconstruct a camouflage image that does little damage to the background structure, but contains the key structure of the foreground. 

\begin{figure}[t]
	\centering
	\small
		\begin{overpic}[width=0.98\linewidth]{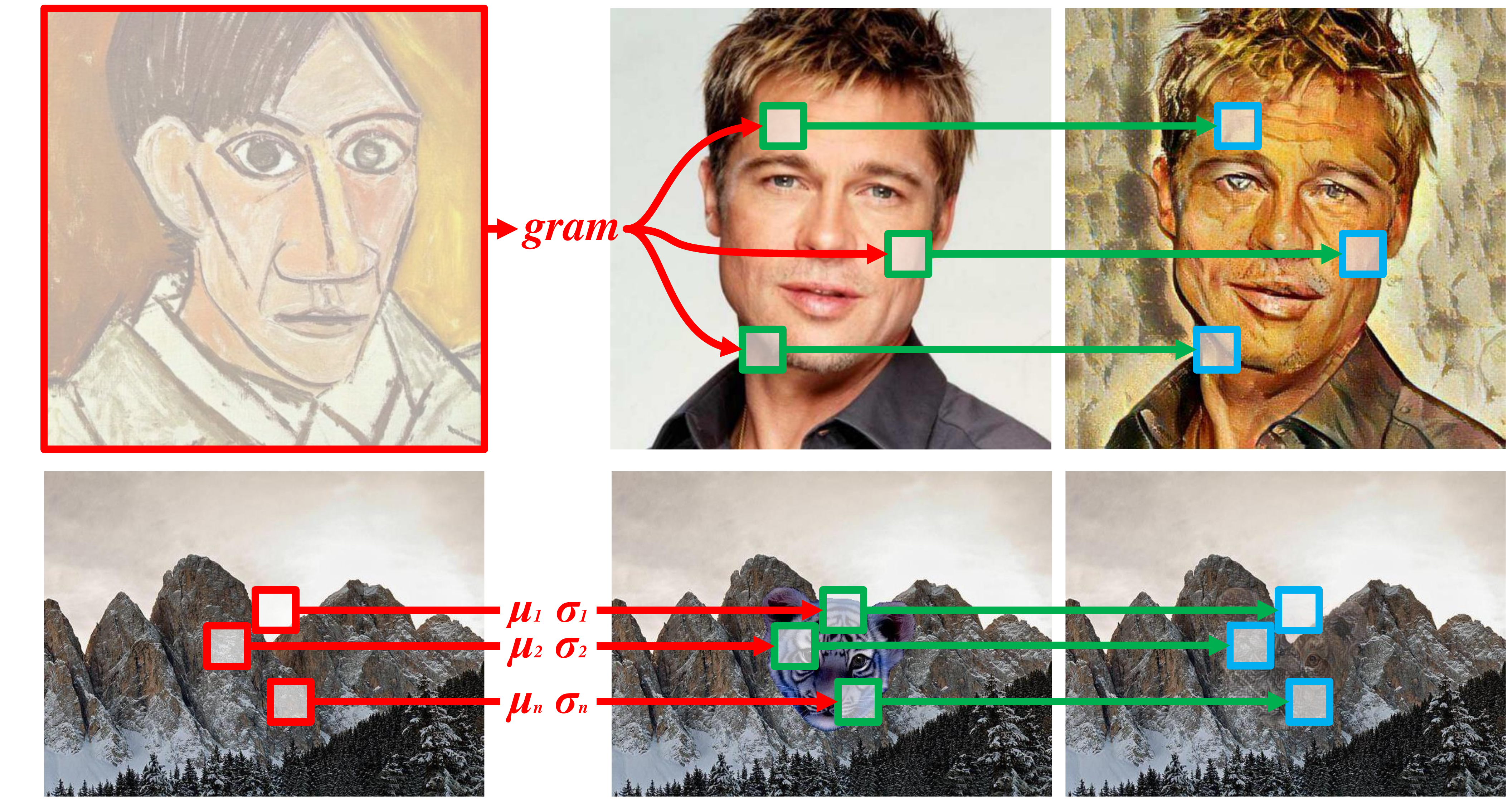}
		\put(-2,39){\textbf{(a)}}
		\put(-2,11){\textbf{(b)}}
		\end{overpic}
	\caption{(a) The style loss used by style transfer \cite{gatys2016image} and Zhang \etal \cite{zhang2020deep}, which extracts global style from the entire style/background images to replace appearance features in the content/foreground images. (b) Schematic of our BPA loss. We compare the local appearance features of each corresponding point in the foreground and background to adapt regions.}
	\label{fig:bpa}
\end{figure}

\textbf{Background Patch Appearance Loss. }In multi-appearance region, different appearances randomly appear at any position. Extracting the global appearance features in this region without distinction will mix different appearances, and fail to recover consistent appearance with the original background. To address this issue, we design a background patch appearance (BPA) loss for this region. For each point in $E(I_o)$, we calculate the mean and variance in a window of size $\omega\times\omega$ and centered on that point, and perform the same operation on $E(I_b)$. BPA loss calculates the difference between them on the entire image: 
\begin{equation}
\label{eq11}
 \begin{split}
  \mathcal{L}_{bpa} = &\sum_{\mathscr{l}=1}^4\sum_i(\left| \mu(E^\mathscr{l}(I_o)_i^{\omega\times\omega}) - \mu(E^\mathscr{l}(I_b)_i^{\omega\times\omega}) \right| \\ 
  &+ \left| \sigma(E^\mathscr{l}(I_o)_i^{\omega\times\omega}) - \sigma(E^\mathscr{l}(I_b)_i^{\omega\times\omega}) \right|), 
 \end{split}
\end{equation}
the definition of each symbol is the same as above.

Similar to Huang \etal \cite{huang2017arbitrary}, BPA loss also obtains the appearance of each region by calculating the mean and variance. But different from the style loss used in style transfer \cite{gatys2016image} and Zhang \etal \cite{zhang2020deep}, BPA loss pays attention to the appearance in a small window of each point. By aligning the local appearance of $I_o$ and $I_b$ at multiple scales, we can effectively divide the different appearances and create a buffer region at their junction, thereby alleviating the camouflage failure caused by the aliasing of multiple appearances. 


\textbf{Total Variational Loss.} As Johnson et al. \cite{johnson2016perceptual} describe, generate approach can not ensure smoothness in the output image. To create a smooth effect on $I_o$, we use total variational loss \cite{perona1990scale,aly2005image} on the pixel-level, it is defined as:
\begin{equation}
\label{eq12}
\begin{aligned}
  \mathcal{L}_{tv} = \sum_{i,j} \left( (I_o^{i,j+1} - I_o^{i,j})^2 + (I_o^{i+1,j} - I_o^{i,j})^2 \right),
\end{aligned}
\end{equation}
where $i$, $j$ represent the index of pixels in the horizontal and vertical directions in $I_o$. 

\section{Experiments}
\label{exp}
\begin{figure}[t]
  \centering
  \includegraphics[width=1\linewidth]{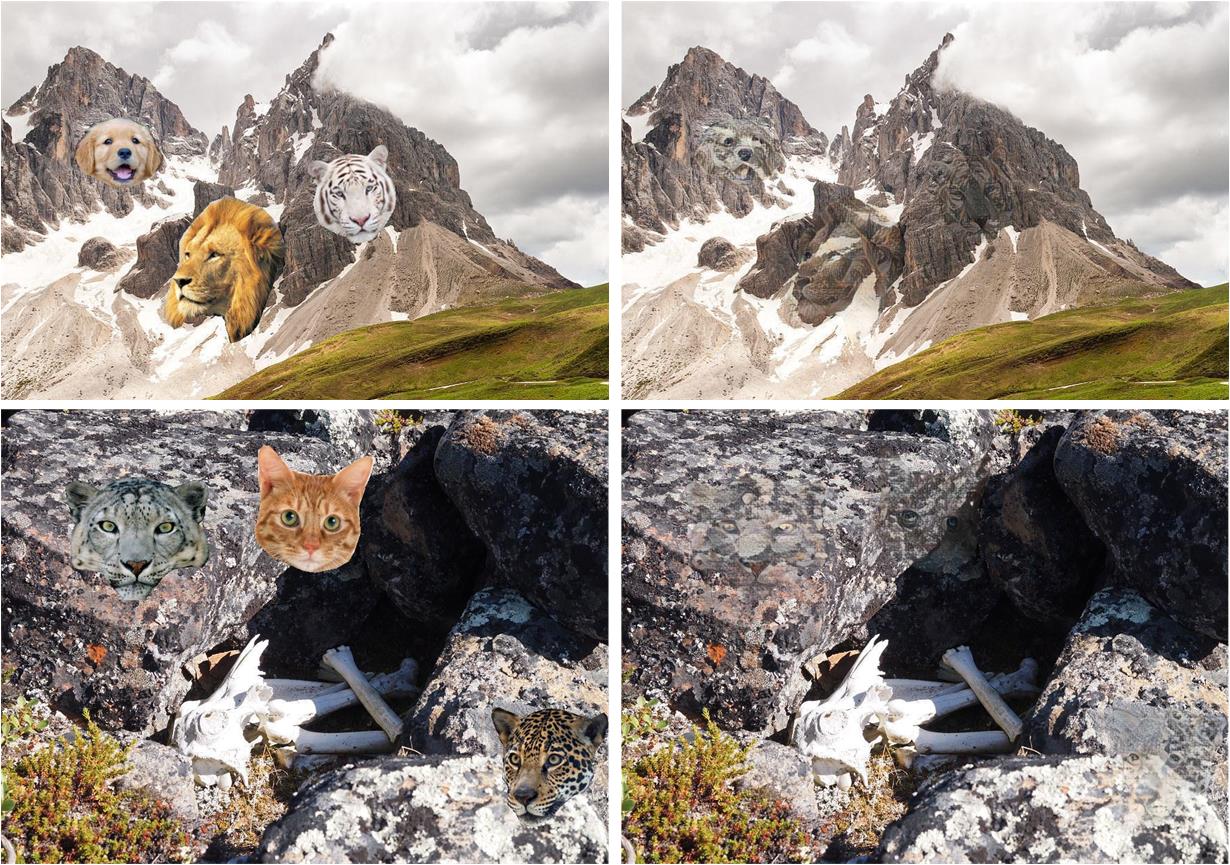}
  \caption{\textbf{Camouflaged images containing multiple foreground objects.} For different types of foreground and background, our method has a good effect.}
  \label{fig:multi_object}
\end{figure}

In this section, we give the detail of our experiment setup. We first compare our approach with existing image blending, style transfer, and camouflage generation approaches in terms of subjective results and user study. We also conduct ablation experiments to prove the influence of the PSF module, each loss component, and different background locations on the generation effect. We additionally show the effect of camouflaging multiple foreground objects simultaneously in Fig. \ref{fig:multi_object}.

\begin{figure*}[t]
  \centering
  \small
		\begin{overpic}[width=1.\linewidth]{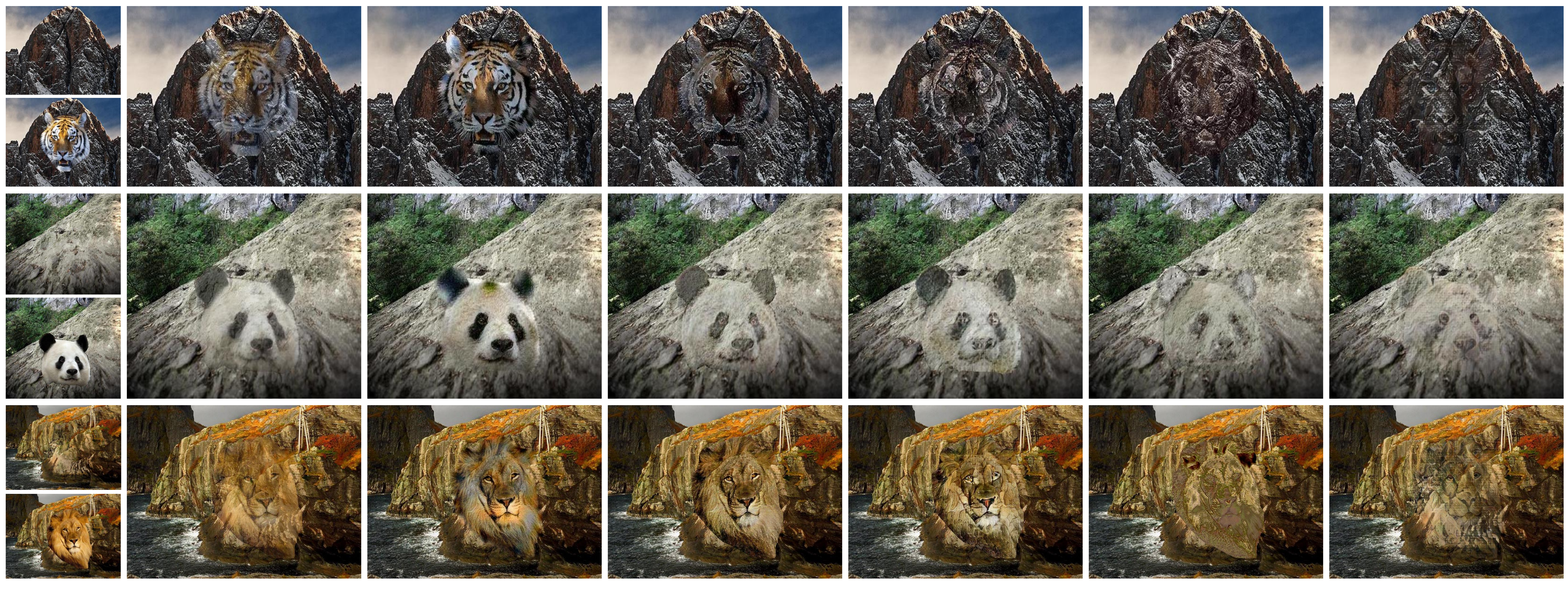}
		\put(1.2, -1){\textbf{B \& E}}
		\put(12.5, -1){\textbf{AB~\cite{perez2003poisson}}}
		\put(28, -1){\textbf{DIB~\cite{zhang2020deepBlend}}}
		\put(44, -1){\textbf{NS~\cite{gatys2016image}}}
		\put(57, -1){\textbf{AdaIN~\cite{huang2017arbitrary}}}
		\put(70, -1){\textbf{Zhang~\etal~\cite{zhang2020deep}}}
		\put(90, -1){\textbf{Ours}}
		\end{overpic}
  \caption{\textbf{Comparison with existing methods in single-appearance regions.} The top of the first column is the background image (B), the bottom is the embedding of the foreground (E), the second to sixth columns are the results generated by other methods, and the last column is our results.}
  \label{fig:single}
\end{figure*}

\begin{figure*}[t]
  \centering
  \small
		\begin{overpic}[width=1.\linewidth]{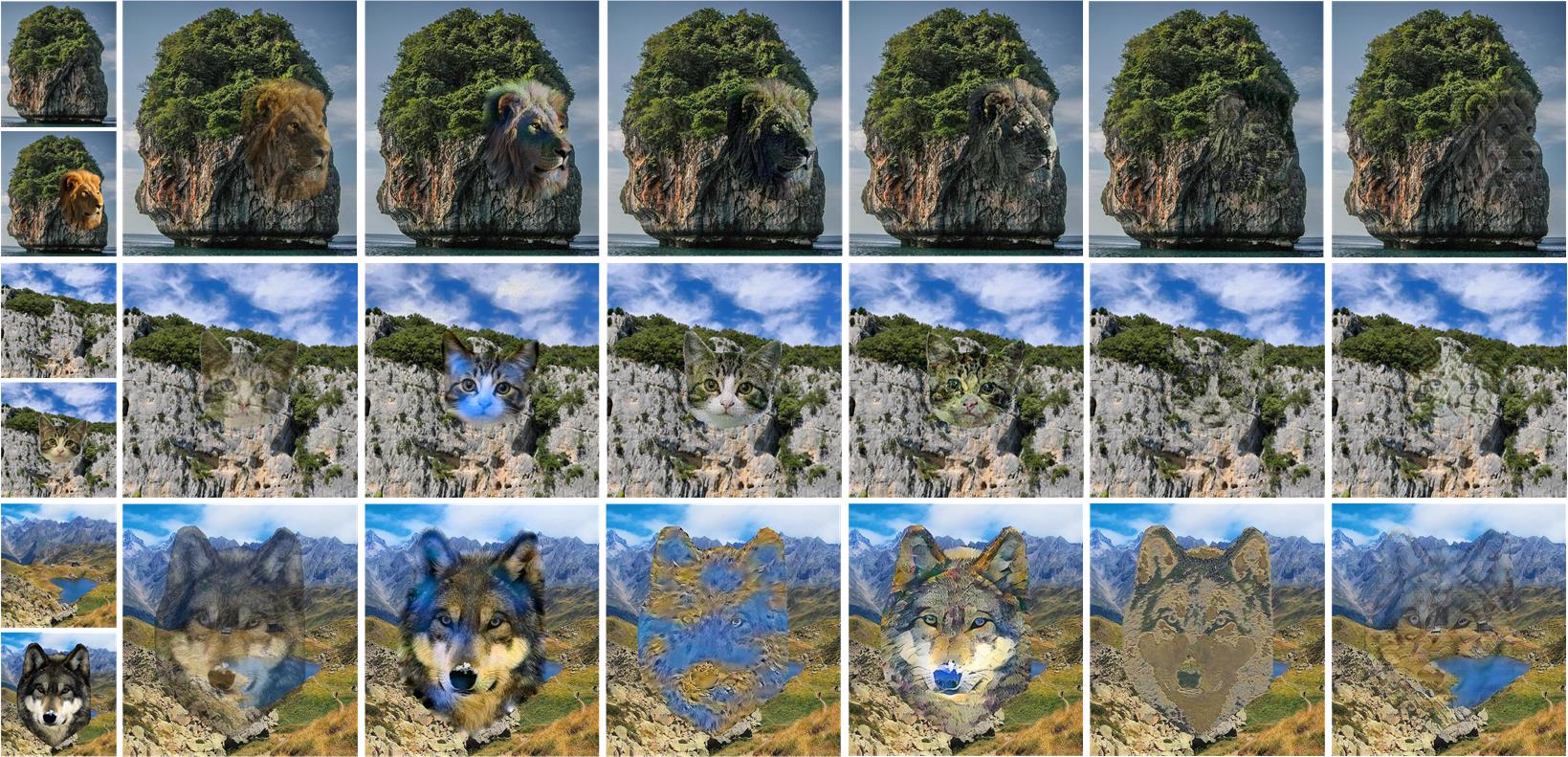}
		\put(1.2, -1.6){\textbf{B \& E}}
		\put(12.5, -1.6){\textbf{AB~\cite{perez2003poisson}}}
		\put(28, -1.6){\textbf{DIB~\cite{zhang2020deepBlend}}}
		\put(44, -1.6){\textbf{NS~\cite{gatys2016image}}}
		\put(57, -1.6){\textbf{AdaIN~\cite{huang2017arbitrary}}}
		\put(70, -1.6){\textbf{Zhang~\etal~\cite{zhang2020deep}}}
		\put(90, -1.6){\textbf{Ours}}
		\end{overpic}
  \caption{\textbf{Comparison with existing methods in multi-appearance regions.} The top of the first column is the background image (B), the bottom is the embedding of the foreground (E), the second to sixth columns are the results generated by other methods, and the last column is our results.}
  \label{fig:multi}
\end{figure*}
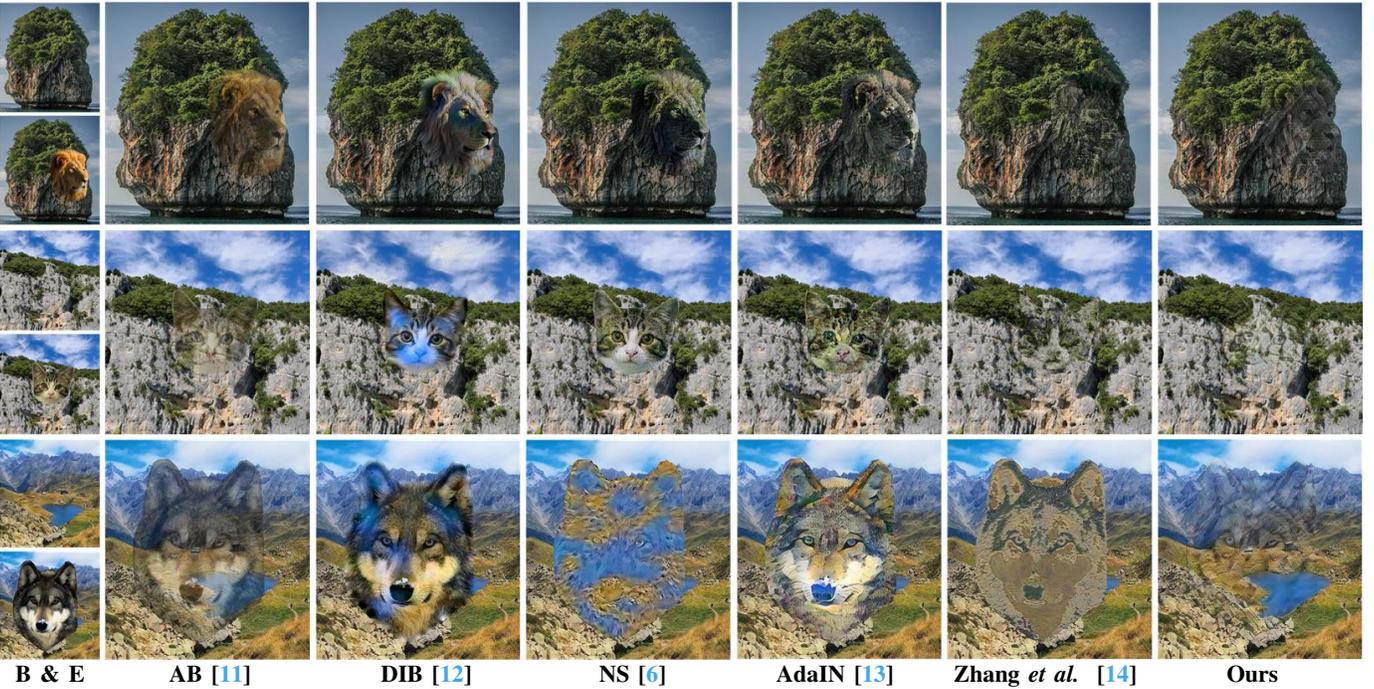

\subsection{Implementation Details}
Our network is trained by 5000 examples in MS-COCO Val 2017 \cite{lin2014microsoft} as foreground images with masks and 4304 examples in Landscape \cite{yu2018land} (cull gray images) as background images. While training, each foreground image is resized to $256 \times 256$, and the background image is resized to $512 \times 512$, then randomly cropped to $256 \times 256$. Our algorithm base on PyTorch \cite{paszke2017automatic}, and use the Adam \cite{kingma2014adam} as optimizer and 8 as batch size. The learning rate is set to $1 \times 10^{-4}$, and the decay rate is set to $5 \times 10^{-5}$. We use the output of {\itshape relu4\_1} in a pre-trained VGG-19 for fusion and IM loss, while the outputs of {\itshape relu1\_1}, {\itshape relu2\_1}, {\itshape relu3\_1} and {\itshape relu4\_1} are chosen for the BPA loss and remove loss. The weight of loss functions is set to $\lambda_{im} = 1.2 \times 10^{4}$, $\lambda_{re} = 1 \times 10^{2}$, $\lambda_{bpa} = 1 \times 10^{2}$, $\lambda_{tv} = 5 \times 10^{-2}$. Our training time on an NVIDIA 1080Ti GPU is 5 hours.

\begin{figure*}[t]
\centering
\begin{overpic}[width=1.\linewidth]{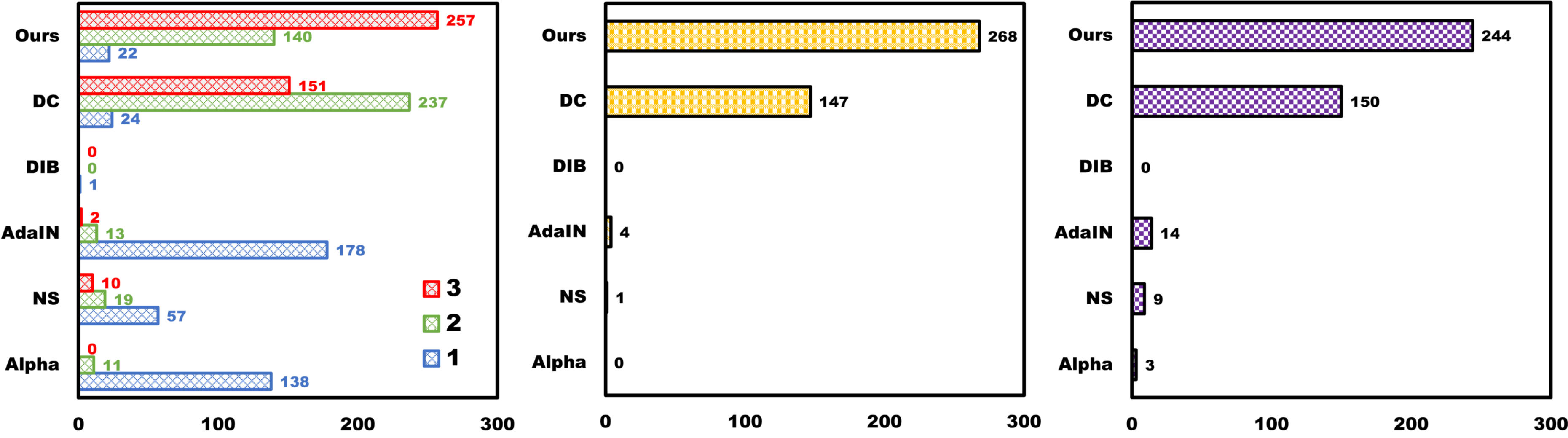}
    \put(0, -1.3){\textbf{\footnotesize{Q\#1: Are the foreground objects hard to be found?}}}
    \put(36, -1.3){\textbf{\footnotesize{Q\#2: Which is the most visually natural result?}}}
    \put(73, -1.3){\textbf{\footnotesize{Q\#3: Which result do you like best?}}}
\end{overpic}
\caption{Ratings for our method and the compared methods on three questions. Our method gets more “3” in \textbf{Q\#1}, which means we have a better camouflage effect. And our method is also considered to be the most visually natural and the favorite.}
\label{fig:user}
\end{figure*}

\begin{table*}[t]
\caption{Speed comparison (in second) with Zhang \etal \cite{zhang2020deep} on the 20 results used in the user study. We record the size of the camouflage regions (not the foreground or background images). Single/Multi (S/M) represent single-appearance/multi-appearance region respectively. Our generation speed is almost independent of size and is two orders of magnitude faster than them.} 
\centering
  \renewcommand{\arraystretch}{1}
  \renewcommand{\tabcolsep}{5pt}
  \small
\resizebox{\textwidth}{22mm}{  
\begin{tabular}{r||c|c|c|c|c|c|c|c|cccc|c}
    \Xhline{2.\arrayrulewidth}
    \hline
    \rowcolor{mygray}
\textbf{Number}                     & \textbf{1}  & \textbf{2}  & \textbf{3}  & \textbf{4}  & \textbf{5}  & \textbf{6}  & \textbf{7}  & \textbf{8}  & \multicolumn{1}{c|}{\textbf{9}} & \multicolumn{1}{c|}{\textbf{10}} & \multicolumn{1}{c|}{\textbf{11}} & \textbf{12} & \textbf{Average Time (S)} \\ 
\Xhline{2.\arrayrulewidth}
\hline
\textbf{Single/Multi}             & S  & S  & S  & S  & S  & S  & S  & S  & \multicolumn{1}{c|}{S} & \multicolumn{1}{c|}{S}  & \multicolumn{1}{c|}{S}  & S  &  S      \\ \hline
\textbf{Size}                       & 
\makecell[c]{$192$ \\ $\times$ \\ $180$} & \begin{tabular}[c]{@{}c@{}}$217$\\ $\times$\\ $238$\end{tabular} & \begin{tabular}[c]{@{}c@{}}$330$\\ $\times$\\ $278$\end{tabular} & \begin{tabular}[c]{@{}c@{}}$179$\\ $\times$\\ $198$\end{tabular} & \begin{tabular}[c]{@{}c@{}}$324$\\ $\times$\\ $389$\end{tabular} & \begin{tabular}[c]{@{}c@{}}$298$\\ $\times$\\ $302$\end{tabular} & \begin{tabular}[c]{@{}c@{}}$258$\\ $\times$\\ $290$\end{tabular} & \begin{tabular}[c]{@{}c@{}}$230$\\ $\times$\\ $288$\end{tabular} & \multicolumn{1}{c|}{\begin{tabular}[c]{@{}c@{}}$235$\\ $\times$\\ $306$\end{tabular}} & \multicolumn{1}{c|}{\begin{tabular}[c]{@{}c@{}}$260$\\ $\times$\\ $287$\end{tabular}} & \multicolumn{1}{c|}{\begin{tabular}[c]{@{}c@{}}$539$\\ $\times$\\ $503$\end{tabular}} &  \multicolumn{1}{c|}{\begin{tabular}[c]{@{}c@{}}$181$\\ $\times$\\ $180$\end{tabular}} & \begin{tabular}[c]{@{}c@{}} $\sim$$270$\\ $\times$\\ $\sim$$287$\end{tabular}  \\ \hline
\textbf{Zhang \etal \cite{zhang2020deep}} 
& $42.40$ & $54.31$ & $87.65$ & $45.33$ & $389.81$ & $264.32$ & $82.62$  & $70.56$            & \multicolumn{1}{c|}{$74.05$}                                                & \multicolumn{1}{c|}{$79.35$}                                                & \multicolumn{1}{c|}{$822.65$}                                               & $453.59$ &  $205.55$       \\ \hline
\textbf{Ours} 
& $0.96$ & $1.10$ & $1.19$ & $1.19$ & $1.00$ & $1.14$ & $1.09$ & $1.06$                       & \multicolumn{1}{c|}{$1.28$}                                                 & \multicolumn{1}{c|}{$1.24$}                                                 & \multicolumn{1}{c|}{$1.39$}                                                 & $1.05$   &  $1.14$   \\ \hline
\Xhline{2.\arrayrulewidth}
\hline
\rowcolor{mygray}
\textbf{Number}                     & \textbf{13} & \textbf{14} & \textbf{15} & \textbf{16} & \textbf{17} & \textbf{18} & \textbf{19} & \textbf{20} & \multicolumn{4}{c|}{\textbf{Average Time (M)}}                                       & \textbf{Total Average Time}    \\ 
\Xhline{2.\arrayrulewidth}
\hline
\textbf{Single/Multi}             & M  & M  & M  & M  & M  & M  & M  & M  & \multicolumn{4}{c|}{M}   &   S \& M       \\ \hline
\textbf{Size}  & \makecell[c]{$254$ \\ $\times$ \\ $255$}  & \makecell[c]{$296$ \\ $\times$ \\ $403$}  & \makecell[c]{$286$ \\ $\times$ \\ $230$}  &
\makecell[c]{$209$ \\ $\times$ \\ $310$}  & \makecell[c]{$242$ \\ $\times$ \\ $207$} & \makecell[c]{$417$ \\ $\times$ \\ $590$}  & \makecell[c]{$316$ \\ $\times$ \\ $357$} & \makecell[c]{$245$ \\ $\times$ \\ $232$}  & 
\multicolumn{4}{c|}{\makecell[c]{ $\sim$$283$ \\ $\times$ \\ $\sim$$323$}}  & \makecell[c]{ $\sim$$275$ \\ $\times$ \\ $\sim$$301$}  \\ \hline
\textbf{Zhang \etal \cite{zhang2020deep}}  & $67.98$ & $364.75$ & $68.05$ & $66.37$ & $56.24$ & $778.18$ & $329.98$ & $59.88$                                                & \multicolumn{4}{c|}{$223.93$}   &   $212.90$ \\ \hline
\textbf{Ours} & $1.05$ & $1.00$ & $1.05$ & $0.95$ & $1.07$  & $1.50$ & $1.00$  & $1.07$       & \multicolumn{4}{c|}{$1.09$} & $1.12$  \\ \hline
\Xhline{2.\arrayrulewidth}
\end{tabular}}
\label{tab:time}
\end{table*}

\subsection{Comparison with Existing Methods}
To show the good generation quality of our approach, we compare the generated results with two image blending methods: alpha blending (AB) \cite{perez2003poisson}, deep image blending (DIB) \cite{zhang2020deepBlend}, two style transfer methods: neural style transfer (NS) \cite{gatys2016image}, AdaIN \cite{huang2017arbitrary}, and a camouflage generation method: deep camouflage image (DC) by Zhang \etal \cite{zhang2020deep}. We choose not to compare with Chu \etal \cite{chu2010camouflage} because their method is a conventional computational camouflage method based on hand-crafted features, and the results of Zhang \etal \cite{zhang2020deep} are more satisfying than theirs. The input of DIB is the entire background image to meet their goal of seamlessly blending the foreground object into the background. For other methods, we use the region corresponding to the smallest square containing the foreground object as background (style) image. We show the comparison results in both single-appearance and multi-appearance regions to prove that our method has the best visual effect in both cases. To show that our method is far faster than DC, we record the time and average time of the two methods on images of different sizes.

{\bfseries Qualitative Comparison.} Fig. \ref{fig:single} and Fig. \ref{fig:multi} show the quality of our approach compared with other approaches in the single-appearance region and multi-appearance region respectively. As shown, image blending methods like AB and DIB cannot make the color and texture of the foreground objects consistent with the background, so they can be easily found. While style transfer methods like NS and AdaIN can transfer the style of the background image to the foreground objects, they retain too many content features and are too abrupt at the edge of the camouflage region, making the objects stand out in most cases. For DC, when the camouflage region has a single appearance (Fig. \ref{fig:single}), it can camouflage the foreground objects to a certain extent, but cannot guarantee that the background structure is not damaged, such as interrupting the continuation of the mountain crack (upper image), or the edge of the camouflage region is not continuous with the rock direction (lower image). In contrast, our method better restores the structure in the background and is more natural. In addition, DC retains too few foreground objects features, which may cause their results to be completely invisible, while the foreground objects in our result can always be found after careful observation. When the camouflage region has multiple appearances (Fig. \ref{fig:multi}), DC either mixes these appearances (middle, lower image) or randomly arranges these appearances (upper image), thus destroying the naturalness of the results and making the foreground objects too standout. In contrast, our method better restores the distribution of different appearances, therefore greatly improving the camouflage effect in multi-appearance regions. 

\begin{figure*}[t]
  \centering
  \small
		\begin{overpic}[width=1.\linewidth]{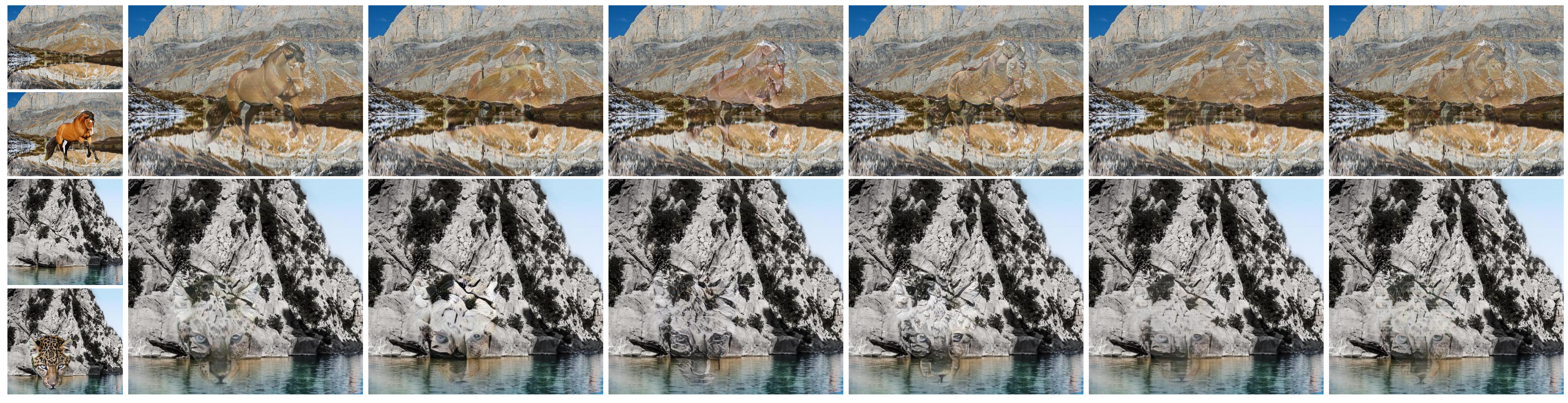}
		\put(-0.5, -1.3){\textbf{(a) B \& E}}
		\put(8.5, -1.3){\textbf{(b) $\mathcal{L}_{im}$ - $\mathcal{L}_{content}$}}
 		\put(25, -1.3){\textbf{(c) $\mathcal{L}_{im}$ - $\mathcal{L}_{leave}$}}
 		\put(40, -1.3){\textbf{(d) $\mathcal{L}_{bpa}$ - $\mathcal{L}_{style}$}}
 		\put(57, -1.3){\textbf{(e) w/o $\mathcal{L}_{re}$}}
 		\put(73, -1.3){\textbf{(f) w/o $\mathcal{L}_{tv}$}}
 		\put(86, -1.3){\textbf{(g) Total Loss}}
		\end{overpic}
  \caption{Ablation study that verifies the role of each component of our loss function. We replace or remove each component in the overall loss individually to examine its impact. ``-'' represents replacing the former with the latter, ``w/o'' represents without. }
  \label{fig:loss}
\end{figure*}

{\bfseries User Study.} The effect of the camouflage image is based on human subjective judgment. Therefore, referring to the previous work on image composition \cite{zhang2020deepBlend, luan2018deep} and camouflage generation \cite{zhang2020deep, chu2010camouflage}, we conduct user studies to quantitatively describe the effect of our results. We randomly select 20 sets of foreground and background images and use different methods to get the corresponding results. We manually select camouflage regions with multiple or single appearances, since random selection yields only a small number of regions with a single appearance. 21 participants are invited to rate the results on three questions: 
\begin{enumerate}

\item [-]
\textbf{Q\#1: Are the foreground objects hard to be found?} (Sort the 3 hardest-to-find results and 3 for the hardest.)

\item [-]
\textbf{Q\#2: Which is the most visually natural result?}

\item [-]
\textbf{Q\#3: Which result do you like best?}

\end{enumerate}

Fig. \ref{fig:user} summarizes the results of the user studies. The three subgraphs correspond to the rating results of the three questions respectively. As shown, our results are considered more difficult to find by users (more number of  ``3'' and rating). Furthermore, our results are also rated as ``the most visually natural'' and ``the most prefer'' by more users. 

{\bfseries Running Time.} Unlike existing methods, our method learns a type of camouflage mapping through data-driven. It only needs one inference to generate camouflage images. Compared with previous camouflage generation methods for optimization iterations, our generation speed is greatly improved. To demonstrate that, we compare our generation time with Zhang \etal \cite{zhang2020deep} on the 20 results in the user study. Since different sizes and whether the camouflaged region has multiple appearances may both affect the generation speed, we record the size and generation time for each image pair and count the average generation time in the single-appearance region and multi-appearance region and total average generation time. Zhang \etal need an additional step of recommending the camouflage region before the camouflage generation, so we only record the time of their camouflage generation step for fairness, which means that they actually take longer than our records. The computing device used by us and Zhang \etal is both an NVIDIA 1080Ti GPU. The comparison results are shown in Table \ref{tab:time}. Our method is two orders of magnitude faster than them, since their method is based on a slow iterative optimization. And the larger camouflage region does not significantly increase our generation time.

\subsection{Ablation Study and Analysis}

{\indent\bfseries Effect of Loss Components.} We provide a detailed analysis of each component in the loss function in FIG. \ref{fig:loss}. The results generated by our full loss are shown in Fig. \ref{fig:loss} (g). In Fig. \ref{fig:loss} (b) and (c), we replace $\mathcal{L}_{im}$ with $\mathcal{L}_{content}$ in style transfer \cite{huang2017arbitrary} and $\mathcal{L}_{leave}$ in Zhang \etal \cite{zhang2020deep}, respectively. It can be found that $\mathcal{L}_{content}$ indiscriminately retain foreground features, making foreground objects too obvious, while $\mathcal{L}_{leave}$ retains too few foreground features except for the most important part in most cases, making the foreground objects difficult to identify. And these problems cannot be solved by simply adjusting the parameters. Our $\mathcal{L}_{im}$ can preserve foreground features in a suitable proportion, so that foreground objects are still identifiable while achieving camouflage. Fig. \ref{fig:loss} (d) shows that using $\mathcal{L}_{style}$ in style transfer \cite{huang2017arbitrary} cannot distinguish the relative positions of different appearances, which causes some appearance features to appear in another location, resulting in an abrupt effect. While $\mathcal{L}_{bpa}$ effectively solves this problem. By introducing $\mathcal{L}_{re}$, we can effectively recover the background structure under the guidance of the saliency map $S$, as shown in Fig. \ref{fig:loss} (e). And by introducing $\mathcal{L}_{tv}$, we can generate a smoother image, as shown in Fig. \ref{fig:loss} (f). 

\begin{figure}[t]
  \centering
	\small
		\begin{overpic}[width=1.\linewidth]{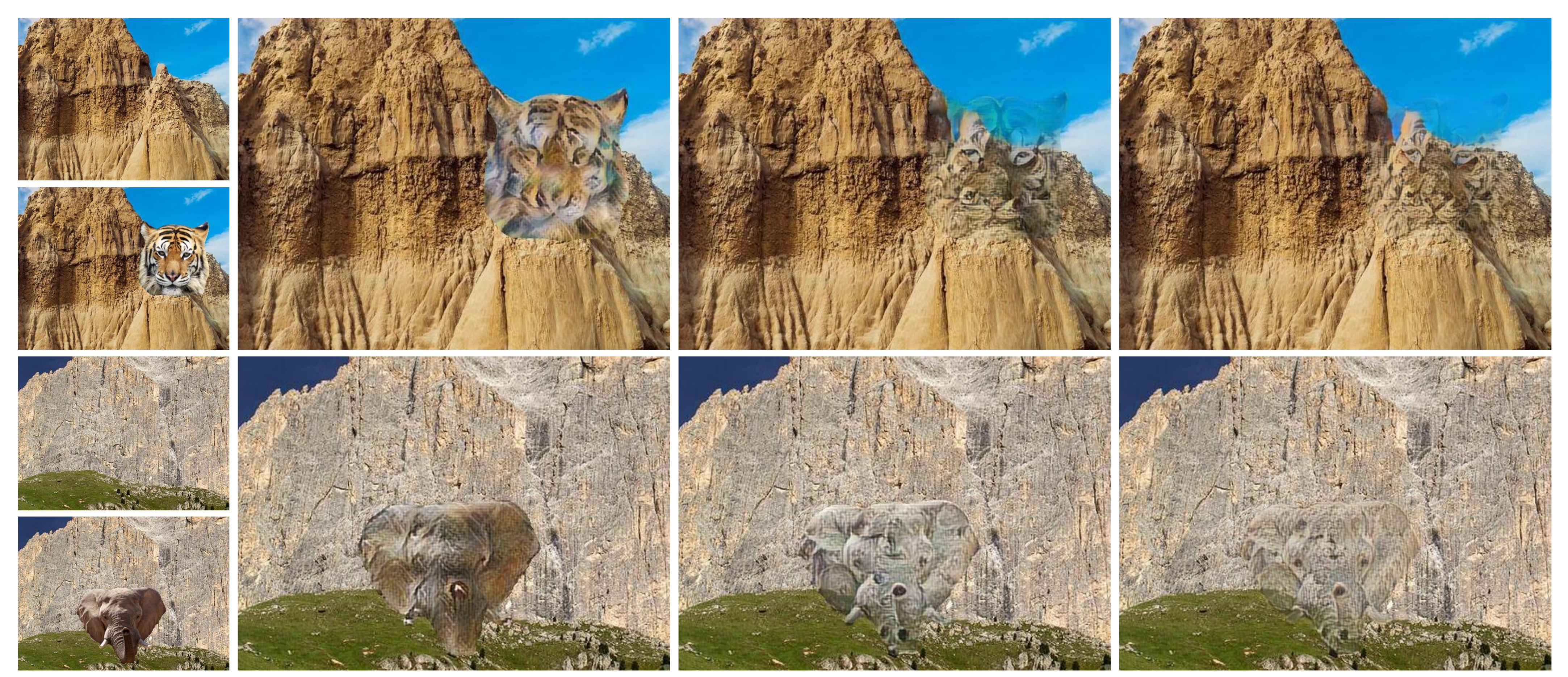}
		\put(0, -3.0){\textbf{(a) B \& E}}
		\put(21, -3.0){\textbf{(b) Adain}}
 		\put(44, -3.0){\textbf{(c) Add + LAda}}
 		\put(78, -3.0){\textbf{(d) PSF}}
		\end{overpic}
  \caption{\textbf{Ablation study that verifies the effect of PSF module.} (b) Using the feature fusion method in AdaIN \cite{huang2017arbitrary}. (c) Summing foreground and background structure features at the channel dimension, and then introducing local appearance features. (d) Our PSF Module.}
  \label{fig:Epsf}
\end{figure}

{\indent\bfseries Effect of PSF.} Here, we investigate the effect of the PSF module in our approach. We design two modules to replace our PSF module: (A) Directly using AdaIN:
\begin{equation}
\label{eq13}
  Module_A(F_f, F_b) = \sigma(F_b)(\frac{F_f - \mu(F_f)}{\sigma(F_f)}) + \mu(F_b),
\end{equation}
which do not consider both structure consistency and appearance consistency. (B) Summing structure features of $F_f$ and $F_b$ at the channel dimension, and introducing local appearance through local AdaIN:
\begin{equation}
\label{eq14}
  Module_B(F_f, F_b) = LAda(\overline{F_f} + \overline{F_b}, F_b),
\end{equation}
which consider the appearance consistency but not the structure consistency. The results generated by A, B and PSF are shown in Fig. \ref{fig:Epsf}. Module A does not preserve the background structure and introduces the mixed appearance features of the entire background image, thus severely destroying the background structure in the camouflage regions and having inconsistent appearances with the surrounding regions such as color and texture (Fig. \ref{fig:Epsf} (b)). Module B introduces the corresponding appearance for foreground objects according to the appearance distribution of the background, but due to the direct summation of the structure features of the foreground and background, the one with the simpler structure is easily completely covered by the other visually (Fig. \ref{fig:Epsf} (c)). Whereas our PSF module generates results with continuous structure and appearance (Fig. \ref{fig:Epsf} (d)).

\begin{figure}[t]
	\centering
	\small
		\begin{overpic}[width=1.\linewidth]{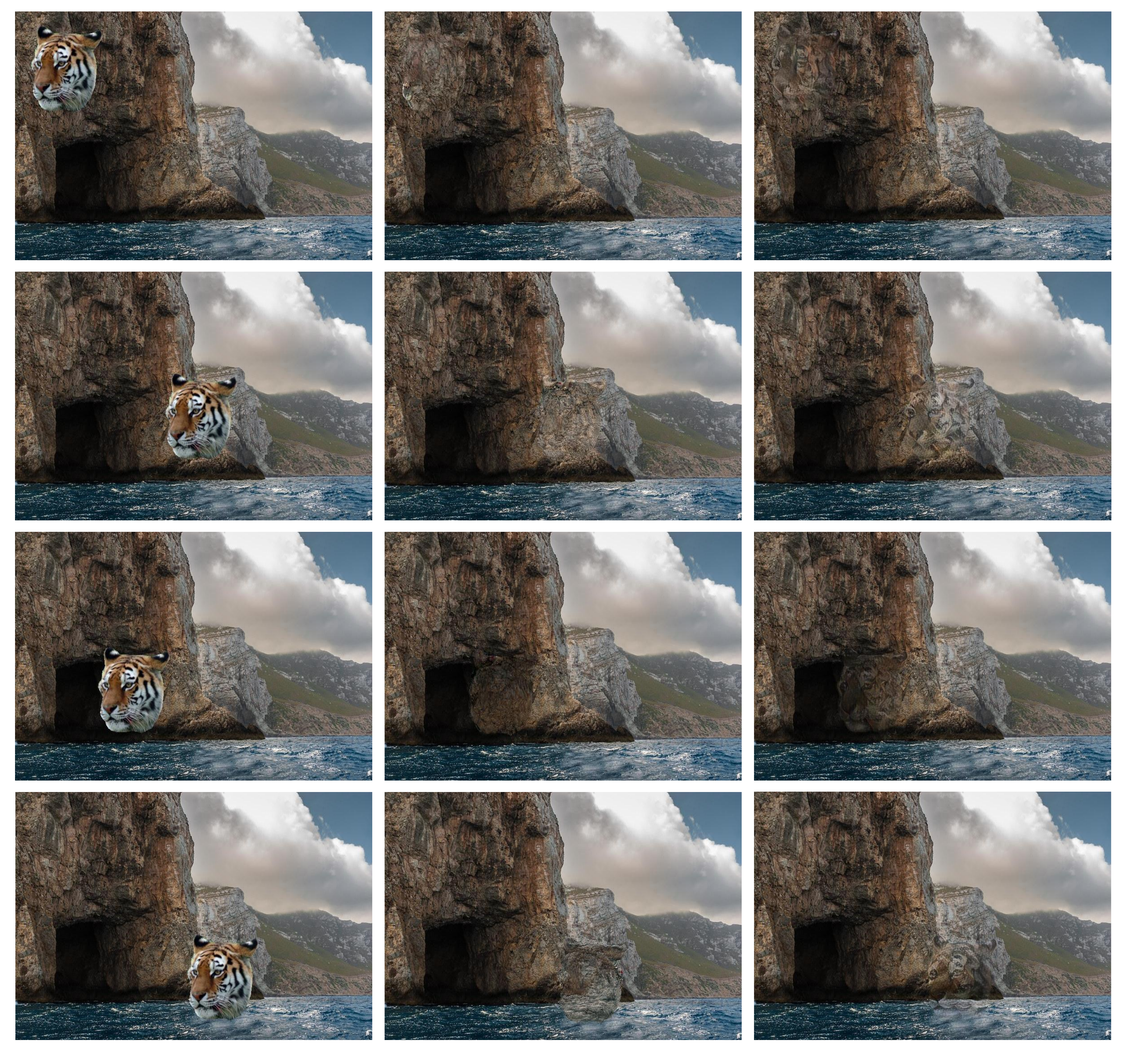}
 		\put(37, -1.5){\textbf{Zhang~\etal~\cite{zhang2020deep}}}
 		\put(79, -1.5){\textbf{Ours}}
		\end{overpic}
	\caption{\textbf{The results of ours and Zhang \etal \cite{zhang2020deep} in different background locations.} They have difficulty generating visually continuous results in multi-appearance regions, and the camouflage effects in single-appearance regions are affected by the different locations. Whereas our results have good visual effects anywhere in the background.}
	\label{fig:location}
\end{figure}

{\indent\bfseries Results for different camouflage location.} Our LCG-Net is able to adapt to camouflage regions with different appearances and has good visual effects at different locations of the background, thus is location-free. Fig. \ref{fig:location} shows the results of ours and Zhang \etal \cite{zhang2020deep} to camouflage the same foreground object at different locations of the same background image. It can be seen that when the camouflage location changes from single-appearance to multi-appearance, Zhang \etal \cite{zhang2020deep} mixes different colors into one color, so it appears abrupt (the second, third, and fourth rows), However, our method has good camouflage effect at different locations of the four images, so it is insensitive to the camouflage location (location-free).  

\section{Application}
\label{app}
As mentioned above, our approach has a fast generation speed and does not limit by background content. Therefore, camouflage generation may have various downstream applications in the fields of art, data augmentation, and information security. Here we envision some potential applications for our approach. We expect our ideas to inspire future exploration and research on camouflage generation. 

\subsection{Camouflage Image Dataset}
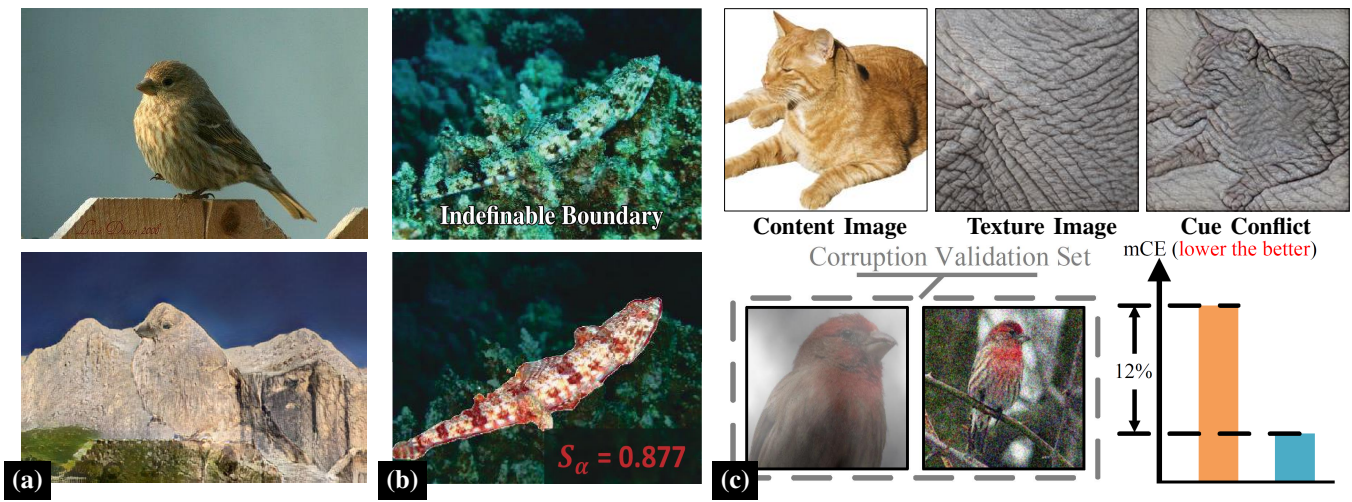
\begin{figure*}[t]
	\centering
	\small
		\begin{overpic}[width=0.98\linewidth]{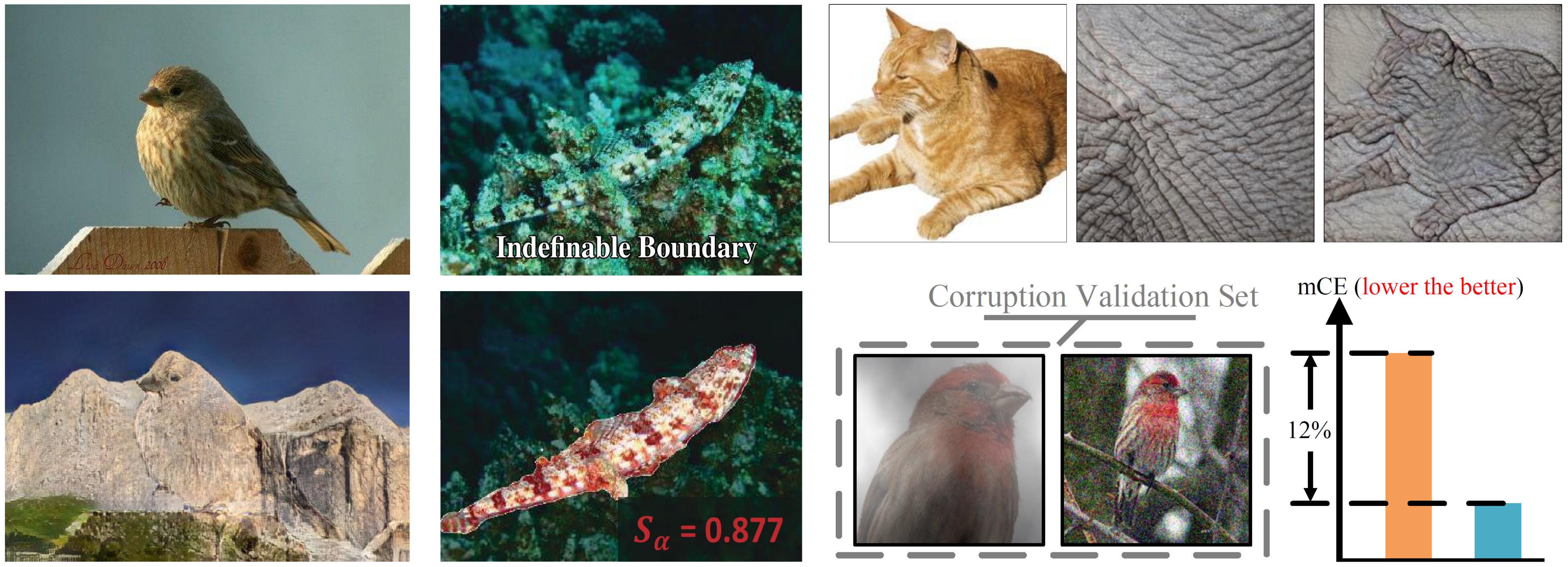}
		\put(-1, 0){\colorbox{black}{{\color{white} \normalsize\textbf{(a)}}}}
		\put(27, 0){\colorbox{black}{{\color{white} \normalsize\textbf{(b)}}}}
		\put(52, 0){\colorbox{black}{{\color{white} \normalsize\textbf{(c)}}}}
		\put(55,19){\textbf{Content Image}}
		\put(71,19){\textbf{Texture Image}}
		\put(87,19){\textbf{Cue Conflict}}
		\end{overpic}
	\caption{\textbf{Camouflage Dataset.} Part (a) A standard example from ImageNet \cite{krizhevsky2012imagenet} and the camouflage example generated with it. We do not embed the foreground objects into the background images since these human visual artifacts have little impact on the deep neural network. Part (b) Concealed Object Detection (COD), the camouflage dataset can assist the training of COD. Part (c) Top: Transfering the style of elephant skin to the cat, Geirhos \etal \cite{geirhos2018imagenet} proves that traditional CNNs are strongly biased towards recognizing textures rather than shapes and classify the stylized result as ``elephant". Bottom: Introducing the camouflage dataset can force CNNs to avoid the features of short-cut and learn the essential features which have the discriminative ability, thereby reducing mean corruption error (mCE) on the corruption validation set.}
	\label{fig:cdata}
\end{figure*}

Although camouflage images are difficult for humans to identify, they can still be seen through careful observation. However, accurately identifying camouflaged objects is more challenging for neural networks than traditional classification/detection, because of the high intrinsic similarities between the targets and non-targets \cite{fan2021concealed}. Therefore, we can use the efficient LCG-Net to quickly generate a series of camouflage datasets to aid training on these tasks, one example generated from ImageNet \cite{krizhevsky2012imagenet} is shown in Fig. \ref{fig:cdata} (a). Since the main role of the camouflage dataset is not for human viewing, we can simply resize the foreground and background images to the same size and then camouflage them. And since our method tends to retain the most salient structure of foreground objects, and the foreground objects are often the most salient in the whole image, we can directly use $I_o$ as the final result without using a mask (and remove $M_d$ in section \ref{method}), which greatly reduces the restriction on the selection of foreground dataset. Camouflage datasets can be used for the following two tasks. 

{\indent\bfseries Concealed Object Detection.}
Concealed object detection (COD) \cite{fan2021concealed} aims to identify objects that are ``seamlessly'' embedded in their surroundings (as shown in Fig. \ref{fig:cdata} (b)). This task has high application potential in medical, agriculture and other fields. However, the high intrinsic similarities between the target object and the background make COD far more challenging than the traditional object detection task. Detection models trained on traditional datasets are difficult to detect camouflaged targets, while the number of natural camouflage images is too small to train a robust model from them. Therefore, we can use the camouflage dataset to assist the training of the COD task, thereby improving the performance of the COD model. 

{\indent\bfseries Improve the Robustness of CNNs.}
Much previous research believes that the most critical feature for humans to recognize objects is shape, and CNNs, like humans, ``implicitly learn representations of shape that reflect human shape perception'' \cite{kubilius2016deep}. However, other studies have pointed out that CNNs are more inclined to the texture features of objects, even if the global shape structure is destroyed, CNNs can still classify images by some certain texture features \cite{gatys2017texture, brendel2019approximating}. However, if the image loses its texture features, it will be difficult for CNNs to recognize \cite{ballester2016performance}. Therefore, the robustness of CNNs in the face of corrupt images still needs to be improved. Geirhos \etal \cite{geirhos2018imagenet} created images with a texture-shape cue conflict by style transfer \cite{gatys2016image} (as shown in the top of Fig. \ref{fig:cdata} (c)), and train a CNN on them to change texture bias towards shape bias. Inspired by them, we can introduce a camouflage dataset, which forces CNNs to avoid the features of short-cut and learn the essential features which have the discriminative ability, thereby improving the robustness against corrupt images (the bottom of Fig. \ref{fig:cdata} (c)). 

\begin{figure}[t]
	\centering
	\small
		\begin{overpic}[width=1.0\linewidth]{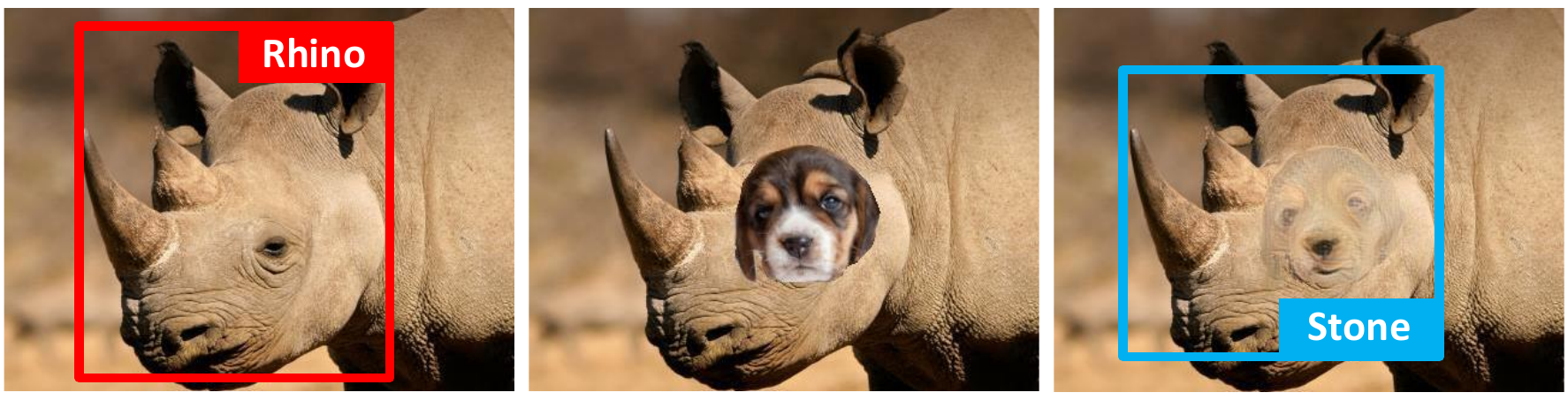}
		\put(14.5,-3){\textbf{(a)}}
		\put(48,-3){\textbf{(b)}}
		\put(81,-3){\textbf{(c)}}
		\end{overpic}
	\caption{\textbf{Adversarial Attack.} (a) Original Image. The detection result is ``rhino''. (b) Embedding the object to be camouflaged. (c) Adversarial Example. It still looks like a ``rhino'', while the detection result is ``stone''.}
	\label{fig:adv}
\end{figure}

\subsection{Adversarial Attack}
With the development of deep neural networks (DNNs), researchers have gradually discovered that a small, carefully designed modification to an input image can cause the network to output wrong results. These modified images are called adversarial examples, and the methods of generating these adversarial examples are called ``Adversarial Attack" \cite{szegedy2013intriguing}. Brown \etal \cite{brown2017adversarial} propose an attack method that affects classifier results by generating an irrelevant patch and placing it anywhere within the field of view of the classifier. Duan \etal \cite{duan2020adversarial} propose an attack method that adds smudges to objects to make the classifier misjudge. Inspired by them, we can camouflage an object in the input image, as shown in Fig. \ref{fig:adv}. The camouflaged object is difficult for humans to find but will cause the classification or detection results to misjudge (\eg identify the ``rhino'' in the background as ``stone''). While for the real-world attack, we can paste carefully crafted camouflage images on the surface of any object to interfere with real-time-demanding tasks such as autonomous driving. Since adversarial attacks and defenses often complement each other, such attack methods may also facilitate the development of adversarial defense techniques. 

\begin{figure}[t]
  \centering
  \includegraphics[width=1\linewidth]{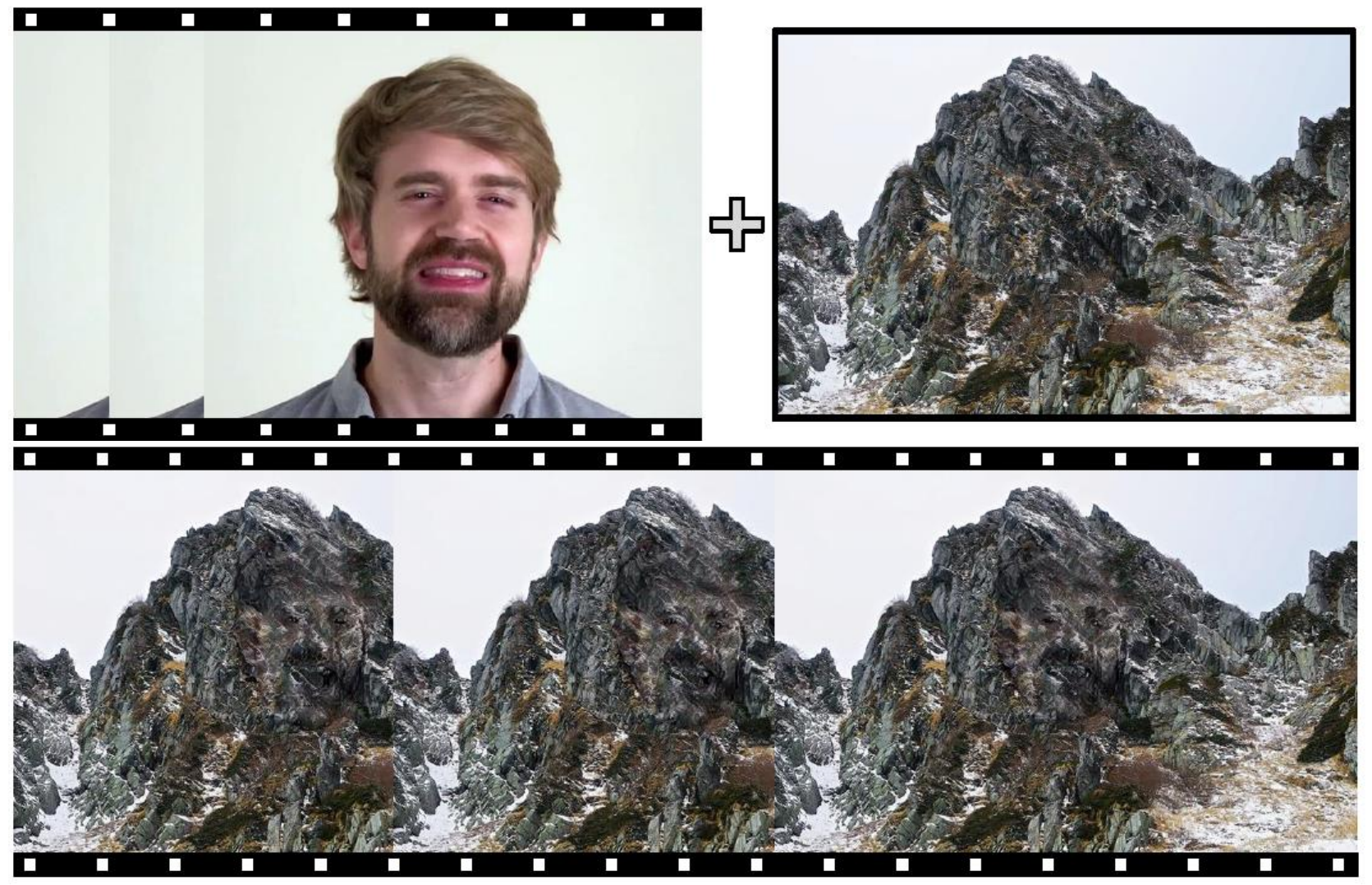}
  \caption{\textbf{Visual Effects.} Camouflage human (or animals) images to mountains (or other landscapes) to create some interesting results, or to use in film and television productions.}
  \label{fig:effect}
\end{figure}

\subsection{Visual Effects}
With the development of photo and short video software, various special effects processing methods emerge one after another. And camouflage images are interesting to humans, so they can be developed into a visual effect. Our LCG-Net can efficiently camouflage foreground objects anywhere in the background image, which allows us to apply it to image and video tasks that contain large numbers of complex backgrounds. Here, we present an easy-to-make and novel demo. Imagine the face of the mountain god emerging on the mountain and speaking, similar special effects can be applied to film and television production or short video entertainment. Specifically, we use LCG-Net to camouflage each frame of the video to the background image and use the video temporal consistency enhancement method proposed by Lai \etal \cite{lai2018learning} to make the processed frames appear more temporally coherent. We show three consecutive frames in Fig. \ref{fig:effect}. 

\begin{figure}[t]
	\centering
	\small
		\begin{overpic}[width=1.0\linewidth]{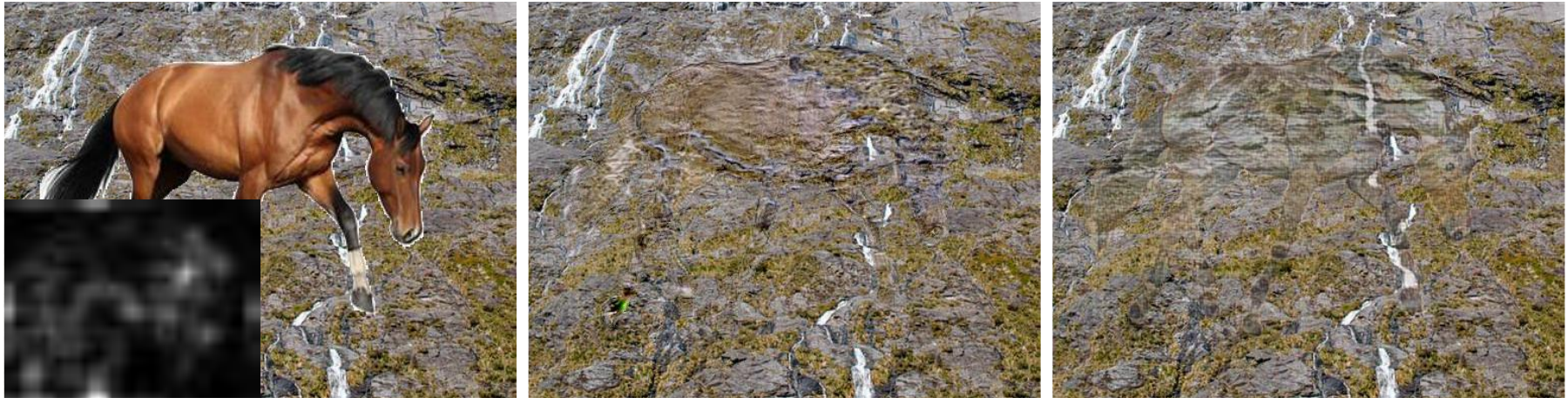}
		\put(6, -2.5){\textbf{Saliency Map}}
		\put(36, -2.5){\textbf{Zhang~\etal~\cite{zhang2020deep}}}
 		\put(79, -2.5){\textbf{Ours}}
		\end{overpic}
	\caption{\textbf{Limitation.} Since the foreground object is too smooth to extract the saliency map, the results of ours and Zhang \etal \cite{zhang2020deep} are both unidentifiable.}
	\label{fig:discriminate}
\end{figure}

\section{Limitation and Discussion}
\label{limitation}
Although our method is able to accomplish the camouflage task in most cases, it still suffers from some limitations. First, we use the saliency map to indicate features that guarantee foreground objects are still identifiable. But in some cases, the part indicated by the saliency map may be different from the part the users wish to retain, which limits the flexibility of our method. A solution is to add the saliency map to the features fed to the decoder during training. After the training, users can modify this item as input to filter the part they want to retain. Second, although both we and Zhang \etal \cite{zhang2020deep} use the saliency map proposed by Hou \etal \cite{hou2007saliency} to indicate features of foreground objects that are sufficiently identifiable, this hand-crafted saliency does not directly equate with identifiability. As shown in Fig \ref{fig:discriminate}, the results generated by the two methods are both difficult to identify because the overall foreground object is too smooth to extract salient parts. To solve this problem, we need to learn which parts of the foreground objects are the most discriminative. A feasible idea is to introduce learning-based methods such as CAM \cite{zhou2016learning} or Grad-CAM \cite{selvaraju2017grad} to replace the saliency map.

\section{Conclusion}
\label{conclusion}
In this paper, we propose a novel end-to-end framework for camouflage image generation, namely location-free camouflage generation network (LCG-Net). Different from the existing methods, we have a more efficient generation speed and use a well-designed position-aligned structure fusion (PSF) module to adaptively combine the high-level features of the foreground and background image to generate a camouflage image that has a good effect even at the multi-appearance regions. Due to the unique advantages of our approach, we envision several potential applications for camouflage generation including camouflage dataset, adversarial attack, and visual effects production. We believe that this flexible and efficient approach can be extended to more interesting and useful tasks. In future work, we plan to study more representational features of foreground objects and perform camouflage on high-resolution images.

\bibliographystyle{IEEEtran}
\bibliography{IEEEtranbib}

\begin{thebibliography}{10}
\providecommand{\url}[1]{#1}
\csname url@samestyle\endcsname
\providecommand{\newblock}{\relax}
\providecommand{\bibinfo}[2]{#2}
\providecommand{\BIBentrySTDinterwordspacing}{\spaceskip=0pt\relax}
\providecommand{\BIBentryALTinterwordstretchfactor}{4}
\providecommand{\BIBentryALTinterwordspacing}{\spaceskip=\fontdimen2\font plus
\BIBentryALTinterwordstretchfactor\fontdimen3\font minus
  \fontdimen4\font\relax}
\providecommand{\BIBforeignlanguage}[2]{{%
\expandafter\ifx\csname l@#1\endcsname\relax
\typeout{** WARNING: IEEEtran.bst: No hyphenation pattern has been}%
\typeout{** loaded for the language `#1'. Using the pattern for}%
\typeout{** the default language instead.}%
\else
\language=\csname l@#1\endcsname
\fi
#2}}
\providecommand{\BIBdecl}{\relax}
\BIBdecl

\bibitem{chu2010camouflage}
H.-K. Chu, W.-H. Hsu, N.~J. Mitra, D.~Cohen-Or, T.-T. Wong, and T.-Y. Lee,
  ``Camouflage images.'' \emph{ACM Trans. Graph.}, vol.~29, no.~4, pp. 51--1,
  2010.

\bibitem{steven2010}
{Steven Michael Gardner}, ``Gardner gallery,''
  \url{http://gardnergallery.com/}, 2010.

\bibitem{kim2019style}
B.-K. Kim, G.~Kim, and S.-Y. Lee, ``Style-controlled synthesis of clothing
  segments for fashion image manipulation,'' \emph{IEEE Transactions on
  Multimedia}, vol.~22, no.~2, pp. 298--310, 2019.

\bibitem{Camouflage20CVPR}
D.-P. Fan, G.-P. Ji, G.~Sun, M.-M. Cheng, J.~Shen, and L.~Shao, ``Camouflaged
  object detection,'' in \emph{IEEE CVPR}, 2020, pp. 2774--2784.

\bibitem{21PAMI-Concealed}
D.-P. Fan, G.-P. Ji, M.-M. Cheng, and L.~Shao, ``Concealed object detection,''
  \emph{IEEE Transactions on Pattern Analysis and Machine Intelligence}, pp.
  1--1, 2021.

\bibitem{gatys2016image}
L.~A. Gatys, A.~S. Ecker, and M.~Bethge, ``Image style transfer using
  convolutional neural networks,'' in \emph{Proceedings of the IEEE conference
  on computer vision and pattern recognition}, 2016, pp. 2414--2423.

\bibitem{gatys2015texture}
L.~Gatys, A.~S. Ecker, and M.~Bethge, ``Texture synthesis using convolutional
  neural networks,'' \emph{Advances in neural information processing systems},
  vol.~28, 2015.

\bibitem{treisman1980feature}
A.~M. Treisman and G.~Gelade, ``A feature-integration theory of attention,''
  \emph{Cognitive psychology}, vol.~12, no.~1, pp. 97--136, 1980.

\bibitem{treisman1988features}
A.~Treisman, ``Features and objects: The fourteenth bartlett memorial
  lecture,'' \emph{The Quarterly Journal of Experimental Psychology Section A},
  vol.~40, no.~2, pp. 201--237, 1988.

\bibitem{wolfe1994guided}
J.~M. Wolfe, ``Guided search 2.0 a revised model of visual search,''
  \emph{Psychonomic bulletin \& review}, vol.~1, no.~2, pp. 202--238, 1994.

\bibitem{perez2003poisson}
P.~P{\'e}rez, M.~Gangnet, and A.~Blake, ``Poisson image editing,'' in \emph{ACM
  SIGGRAPH 2003 Papers}, 2003, pp. 313--318.

\bibitem{zhang2020deepBlend}
L.~Zhang, T.~Wen, and J.~Shi, ``Deep image blending,'' in \emph{Proceedings of
  the IEEE/CVF Winter Conference on Applications of Computer Vision}, 2020, pp.
  231--240.

\bibitem{huang2017arbitrary}
X.~Huang and S.~Belongie, ``Arbitrary style transfer in real-time with adaptive
  instance normalization,'' in \emph{Proceedings of the IEEE International
  Conference on Computer Vision}, 2017, pp. 1501--1510.

\bibitem{zhang2020deep}
Q.~Zhang, G.~Yin, Y.~Nie, and W.-S. Zheng, ``Deep camouflage images,'' in
  \emph{Proceedings of the AAAI Conference on Artificial Intelligence},
  vol.~34, no.~07, 2020, pp. 12\,845--12\,852.

\bibitem{heeger1995pyramid}
D.~J. Heeger and J.~R. Bergen, ``Pyramid-based texture analysis/synthesis,'' in
  \emph{Proceedings of the 22nd annual conference on Computer graphics and
  interactive techniques}, 1995, pp. 229--238.

\bibitem{efros1999texture}
A.~A. Efros and T.~K. Leung, ``Texture synthesis by non-parametric sampling,''
  in \emph{Proceedings of the seventh IEEE international conference on computer
  vision}, vol.~2.\hskip 1em plus 0.5em minus 0.4em\relax IEEE, 1999, pp.
  1033--1038.

\bibitem{efros2001image}
A.~A. Efros and W.~T. Freeman, ``Image quilting for texture synthesis and
  transfer,'' in \emph{Proceedings of the 28th annual conference on Computer
  graphics and interactive techniques}, 2001, pp. 341--346.

\bibitem{lockerman2016multi}
Y.~D. Lockerman, B.~Sauvage, R.~All{\`e}gre, J.-M. Dischler, J.~Dorsey, and
  H.~E. Rushmeier, ``Multi-scale label-map extraction for texture synthesis.''
  \emph{ACM Trans. Graph.}, vol.~35, no.~4, pp. 140--1, 2016.

\bibitem{li2018non}
J.~Li, Y.~Xiang, J.~Hou, and D.~Xu, ``Non-local texture optimization with
  wasserstein regularization under convolutional neural network,'' \emph{IEEE
  Transactions on Multimedia}, vol.~21, no.~6, pp. 1437--1449, 2018.

\bibitem{frigo2016split}
O.~Frigo, N.~Sabater, J.~Delon, and P.~Hellier, ``Split and match:
  Example-based adaptive patch sampling for unsupervised style transfer,'' in
  \emph{Proceedings of the IEEE Conference on Computer Vision and Pattern
  Recognition}, 2016, pp. 553--561.

\bibitem{johnson2016perceptual}
J.~Johnson, A.~Alahi, and L.~Fei-Fei, ``Perceptual losses for real-time style
  transfer and super-resolution,'' in \emph{European conference on computer
  vision}.\hskip 1em plus 0.5em minus 0.4em\relax Springer, 2016, pp. 694--711.

\bibitem{ulyanov2016texture}
D.~Ulyanov, V.~Lebedev, A.~Vedaldi, and V.~S. Lempitsky, ``Texture networks:
  Feed-forward synthesis of textures and stylized images.'' in \emph{ICML},
  vol.~1, no.~2, 2016, p.~4.

\bibitem{li2016precomputed}
C.~Li and M.~Wand, ``Precomputed real-time texture synthesis with markovian
  generative adversarial networks,'' in \emph{European conference on computer
  vision}.\hskip 1em plus 0.5em minus 0.4em\relax Springer, 2016, pp. 702--716.

\bibitem{ulyanov2017improved}
D.~Ulyanov, A.~Vedaldi, and V.~Lempitsky, ``Improved texture networks:
  Maximizing quality and diversity in feed-forward stylization and texture
  synthesis,'' in \emph{Proceedings of the IEEE Conference on Computer Vision
  and Pattern Recognition}, 2017, pp. 6924--6932.

\bibitem{wang2017multimodal}
X.~Wang, G.~Oxholm, D.~Zhang, and Y.-F. Wang, ``Multimodal transfer: A
  hierarchical deep convolutional neural network for fast artistic style
  transfer,'' in \emph{Proceedings of the IEEE Conference on Computer Vision
  and Pattern Recognition}, 2017, pp. 5239--5247.

\bibitem{li2017universal}
Y.~Li, C.~Fang, J.~Yang, Z.~Wang, X.~Lu, and M.-H. Yang, ``Universal style
  transfer via feature transforms,'' \emph{arXiv preprint arXiv:1705.08086},
  2017.

\bibitem{jing2020dynamic}
Y.~Jing, X.~Liu, Y.~Ding, X.~Wang, E.~Ding, M.~Song, and S.~Wen, ``Dynamic
  instance normalization for arbitrary style transfer,'' in \emph{Proceedings
  of the AAAI Conference on Artificial Intelligence}, vol.~34, no.~04, 2020,
  pp. 4369--4376.

\bibitem{li2018learning}
X.~Li, S.~Liu, J.~Kautz, and M.-H. Yang, ``Learning linear transformations for
  fast arbitrary style transfer,'' \emph{arXiv preprint arXiv:1808.04537},
  2018.

\bibitem{deng2020arbitrary}
Y.~Deng, F.~Tang, W.~Dong, H.~Huang, C.~Ma, and C.~Xu, ``Arbitrary video style
  transfer via multi-channel correlation,'' \emph{arXiv preprint
  arXiv:2009.08003}, 2020.

\bibitem{chen2016fast}
T.~Q. Chen and M.~Schmidt, ``Fast patch-based style transfer of arbitrary
  style,'' \emph{arXiv preprint arXiv:1612.04337}, 2016.

\bibitem{sheng2018avatar}
L.~Sheng, Z.~Lin, J.~Shao, and X.~Wang, ``Avatar-net: Multi-scale zero-shot
  style transfer by feature decoration,'' in \emph{Proceedings of the IEEE
  Conference on Computer Vision and Pattern Recognition}, 2018, pp. 8242--8250.

\bibitem{park2019arbitrary}
D.~Y. Park and K.~H. Lee, ``Arbitrary style transfer with style-attentional
  networks,'' in \emph{Proceedings of the IEEE/CVF Conference on Computer
  Vision and Pattern Recognition}, 2019, pp. 5880--5888.

\bibitem{liu2021adaattn}
S.~Liu, T.~Lin, D.~He, F.~Li, M.~Wang, X.~Li, Z.~Sun, Q.~Li, and E.~Ding,
  ``Adaattn: Revisit attention mechanism in arbitrary neural style transfer,''
  in \emph{Proceedings of the IEEE/CVF International Conference on Computer
  Vision}, 2021, pp. 6649--6658.

\bibitem{virtusio2021neural}
J.~J. Virtusio, J.~J.~M. Ople, D.~S. Tan, M.~Tanveer, N.~Kumar, and K.-L. Hua,
  ``Neural style palette: A multimodal and interactive style transfer from a
  single style image,'' \emph{IEEE Transactions on Multimedia}, vol.~23, pp.
  2245--2258, 2021.

\bibitem{gupta2017characterizing}
A.~Gupta, J.~Johnson, A.~Alahi, and L.~Fei-Fei, ``Characterizing and improving
  stability in neural style transfer,'' in \emph{Proceedings of the IEEE
  International Conference on Computer Vision}, 2017, pp. 4067--4076.

\bibitem{chen2017coherent}
D.~Chen, J.~Liao, L.~Yuan, N.~Yu, and G.~Hua, ``Coherent online video style
  transfer,'' in \emph{Proceedings of the IEEE International Conference on
  Computer Vision}, 2017, pp. 1105--1114.

\bibitem{huang2017real}
H.~Huang, H.~Wang, W.~Luo, L.~Ma, W.~Jiang, X.~Zhu, Z.~Li, and W.~Liu,
  ``Real-time neural style transfer for videos,'' in \emph{Proceedings of the
  IEEE Conference on Computer Vision and Pattern Recognition}, 2017, pp.
  783--791.

\bibitem{agarwala2004interactive}
A.~Agarwala, M.~Dontcheva, M.~Agrawala, S.~Drucker, A.~Colburn, B.~Curless,
  D.~Salesin, and M.~Cohen, ``Interactive digital photomontage,'' in \emph{ACM
  SIGGRAPH 2004 Papers}, 2004, pp. 294--302.

\bibitem{jia2006drag}
J.~Jia, J.~Sun, C.-K. Tang, and H.-Y. Shum, ``Drag-and-drop pasting,''
  \emph{ACM Transactions on graphics (TOG)}, vol.~25, no.~3, pp. 631--637,
  2006.

\bibitem{levin2004seamless}
A.~Levin, A.~Zomet, S.~Peleg, and Y.~Weiss, ``Seamless image stitching in the
  gradient domain,'' in \emph{European conference on computer vision}.\hskip
  1em plus 0.5em minus 0.4em\relax Springer, 2004, pp. 377--389.

\bibitem{uyttendaele2001eliminating}
M.~Uyttendaele, A.~Eden, and R.~Skeliski, ``Eliminating ghosting and exposure
  artifacts in image mosaics,'' in \emph{Proceedings of the 2001 IEEE Computer
  Society Conference on Computer Vision and Pattern Recognition. CVPR 2001},
  vol.~2.\hskip 1em plus 0.5em minus 0.4em\relax IEEE, 2001, pp. II--II.

\bibitem{fattal2002gradient}
R.~Fattal, D.~Lischinski, and M.~Werman, ``Gradient domain high dynamic range
  compression,'' in \emph{Proceedings of the 29th annual conference on Computer
  graphics and interactive techniques}, 2002, pp. 249--256.

\bibitem{kazhdan2008streaming}
M.~Kazhdan and H.~Hoppe, ``Streaming multigrid for gradient-domain operations
  on large images,'' \emph{ACM Transactions on graphics (TOG)}, vol.~27, no.~3,
  pp. 1--10, 2008.

\bibitem{luan2018deep}
F.~Luan, S.~Paris, E.~Shechtman, and K.~Bala, ``Deep painterly harmonization,''
  in \emph{Computer graphics forum}, vol.~37, no.~4.\hskip 1em plus 0.5em minus
  0.4em\relax Wiley Online Library, 2018, pp. 95--106.

\bibitem{xiao2019multi}
B.~Xiao, G.~Ou, H.~Tang, X.~Bi, and W.~Li, ``Multi-focus image fusion by
  hessian matrix based decomposition,'' \emph{IEEE Transactions on Multimedia},
  vol.~22, no.~2, pp. 285--297, 2019.

\bibitem{zhang2016reversible}
W.~Zhang, H.~Wang, D.~Hou, and N.~Yu, ``Reversible data hiding in encrypted
  images by reversible image transformation,'' \emph{IEEE Transactions on
  multimedia}, vol.~18, no.~8, pp. 1469--1479, 2016.

\bibitem{tankus2001convexity}
A.~Tankus and Y.~Yeshurun, ``Convexity-based visual camouflage breaking,''
  \emph{Computer Vision and Image Understanding}, vol.~82, no.~3, pp. 208--237,
  2001.

\bibitem{reynolds2011interactive}
C.~Reynolds, ``Interactive evolution of camouflage,'' \emph{Artificial life},
  vol.~17, no.~2, pp. 123--136, 2011.

\bibitem{fan2021concealed}
D.-P. Fan, G.-P. Ji, M.-M. Cheng, and L.~Shao, ``Concealed object detection,''
  \emph{arXiv preprint arXiv:2102.10274}, 2021.

\bibitem{mei2021camouflaged}
H.~Mei, G.-P. Ji, Z.~Wei, X.~Yang, X.~Wei, and D.-P. Fan, ``Camouflaged object
  segmentation with distraction mining,'' in \emph{Proceedings of the IEEE/CVF
  Conference on Computer Vision and Pattern Recognition}, 2021, pp. 8772--8781.

\bibitem{zhai2021mutual}
Q.~Zhai, X.~Li, F.~Yang, C.~Chen, H.~Cheng, and D.-P. Fan, ``Mutual graph
  learning for camouflaged object detection,'' in \emph{Proceedings of the
  IEEE/CVF Conference on Computer Vision and Pattern Recognition}, 2021, pp.
  12\,997--13\,007.

\bibitem{simonyan2014very}
K.~Simonyan and A.~Zisserman, ``Very deep convolutional networks for
  large-scale image recognition,'' \emph{arXiv preprint arXiv:1409.1556}, 2014.

\bibitem{dumoulin2016learned}
V.~Dumoulin, J.~Shlens, and M.~Kudlur, ``A learned representation for artistic
  style,'' \emph{arXiv preprint arXiv:1610.07629}, 2016.

\bibitem{hou2007saliency}
X.~Hou and L.~Zhang, ``Saliency detection: A spectral residual approach,'' in
  \emph{2007 IEEE Conference on computer vision and pattern recognition}.\hskip
  1em plus 0.5em minus 0.4em\relax Ieee, 2007, pp. 1--8.

\bibitem{perona1990scale}
P.~Perona and J.~Malik, ``Scale-space and edge detection using anisotropic
  diffusion,'' \emph{IEEE Transactions on pattern analysis and machine
  intelligence}, vol.~12, no.~7, pp. 629--639, 1990.

\bibitem{aly2005image}
H.~A. Aly and E.~Dubois, ``Image up-sampling using total-variation
  regularization with a new observation model,'' \emph{IEEE Transactions on
  Image Processing}, vol.~14, no.~10, pp. 1647--1659, 2005.

\bibitem{lin2014microsoft}
T.-Y. Lin, M.~Maire, S.~Belongie, J.~Hays, P.~Perona, D.~Ramanan,
  P.~Doll{\'a}r, and C.~L. Zitnick, ``Microsoft coco: Common objects in
  context,'' in \emph{European conference on computer vision}.\hskip 1em plus
  0.5em minus 0.4em\relax Springer, 2014, pp. 740--755.

\bibitem{yu2018land}
\BIBentryALTinterwordspacing
W.~Yu. (2018, Jun.) Landscape-dataset. [Online]. Available:
  \url{https://github.com/yuweiming70/Landscape-Dataset}
\BIBentrySTDinterwordspacing

\bibitem{paszke2017automatic}
A.~Paszke, S.~Gross, S.~Chintala, G.~Chanan, E.~Yang, Z.~DeVito, Z.~Lin,
  A.~Desmaison, L.~Antiga, and A.~Lerer, ``Automatic differentiation in
  pytorch,'' 2017.

\bibitem{kingma2014adam}
D.~P. Kingma and J.~Ba, ``Adam: A method for stochastic optimization,''
  \emph{arXiv preprint arXiv:1412.6980}, 2014.

\bibitem{krizhevsky2012imagenet}
A.~Krizhevsky, I.~Sutskever, and G.~E. Hinton, ``Imagenet classification with
  deep convolutional neural networks,'' \emph{Advances in neural information
  processing systems}, vol.~25, pp. 1097--1105, 2012.

\bibitem{geirhos2018imagenet}
R.~Geirhos, P.~Rubisch, C.~Michaelis, M.~Bethge, F.~A. Wichmann, and
  W.~Brendel, ``Imagenet-trained cnns are biased towards texture; increasing
  shape bias improves accuracy and robustness,'' \emph{arXiv preprint
  arXiv:1811.12231}, 2018.

\bibitem{kubilius2016deep}
J.~Kubilius, S.~Bracci, and H.~P. Op~de Beeck, ``Deep neural networks as a
  computational model for human shape sensitivity,'' \emph{PLoS computational
  biology}, vol.~12, no.~4, p. e1004896, 2016.

\bibitem{gatys2017texture}
L.~A. Gatys, A.~S. Ecker, and M.~Bethge, ``Texture and art with deep neural
  networks,'' \emph{Current opinion in neurobiology}, vol.~46, pp. 178--186,
  2017.

\bibitem{brendel2019approximating}
W.~Brendel and M.~Bethge, ``Approximating cnns with bag-of-local-features
  models works surprisingly well on imagenet,'' \emph{arXiv preprint
  arXiv:1904.00760}, 2019.

\bibitem{ballester2016performance}
P.~Ballester and R.~Araujo, ``On the performance of googlenet and alexnet
  applied to sketches,'' in \emph{Proceedings of the AAAI Conference on
  Artificial Intelligence}, vol.~30, no.~1, 2016.

\bibitem{szegedy2013intriguing}
C.~Szegedy, W.~Zaremba, I.~Sutskever, J.~Bruna, D.~Erhan, I.~Goodfellow, and
  R.~Fergus, ``Intriguing properties of neural networks,'' \emph{arXiv preprint
  arXiv:1312.6199}, 2013.

\bibitem{brown2017adversarial}
T.~B. Brown, D.~Man{\'e}, A.~Roy, M.~Abadi, and J.~Gilmer, ``Adversarial
  patch,'' \emph{arXiv preprint arXiv:1712.09665}, 2017.

\bibitem{duan2020adversarial}
R.~Duan, X.~Ma, Y.~Wang, J.~Bailey, A.~K. Qin, and Y.~Yang, ``Adversarial
  camouflage: Hiding physical-world attacks with natural styles,'' in
  \emph{Proceedings of the IEEE/CVF conference on computer vision and pattern
  recognition}, 2020, pp. 1000--1008.

\bibitem{lai2018learning}
W.-S. Lai, J.-B. Huang, O.~Wang, E.~Shechtman, E.~Yumer, and M.-H. Yang,
  ``Learning blind video temporal consistency,'' in \emph{Proceedings of the
  European conference on computer vision (ECCV)}, 2018, pp. 170--185.

\bibitem{zhou2016learning}
B.~Zhou, A.~Khosla, A.~Lapedriza, A.~Oliva, and A.~Torralba, ``Learning deep
  features for discriminative localization,'' in \emph{Proceedings of the IEEE
  conference on computer vision and pattern recognition}, 2016, pp. 2921--2929.

\bibitem{selvaraju2017grad}
R.~R. Selvaraju, M.~Cogswell, A.~Das, R.~Vedantam, D.~Parikh, and D.~Batra,
  ``Grad-cam: Visual explanations from deep networks via gradient-based
  localization,'' in \emph{Proceedings of the IEEE international conference on
  computer vision}, 2017, pp. 618--626.

\end{thebibliography}

\end{document}